\documentclass[letterpaper]{article} % DO NOT CHANGE THIS
\usepackage{aaai2027}  % DO NOT CHANGE THIS
\usepackage[hyphens]{url}  % DO NOT CHANGE THIS
\usepackage{graphicx} % DO NOT CHANGE THIS
\usepackage{natbib}  % DO NOT CHANGE THIS AND DO NOT ADD ANY OPTIONS TO IT
\usepackage{caption} % DO NOT CHANGE THIS AND DO NOT ADD ANY OPTIONS TO IT
\nocopyright % arXiv preprint: not an accepted AAAI paper, so no AAAI copyright line

\usepackage{amsmath}
\usepackage{amssymb}
\usepackage{amsfonts}
\usepackage{amsthm}
\usepackage{booktabs}
\usepackage{algorithm}
\usepackage{algorithmic}

\newtheorem{theorem}{Theorem}
\newtheorem{proposition}{Proposition}
\newtheorem{corollary}{Corollary}
\newtheorem{lemma}{Lemma}
\newtheorem{definition}{Definition}

\newtheorem{remark}{Remark}

\title{Certified Comparative Statics for Conservation-Law Equilibrium Layers, with an Application to RNA Interference}

\author{
    Shadi Khodagholi\textsuperscript{1},\quad
    Sara Karimi\textsuperscript{2},\quad
    Niloofar Yousefi\textsuperscript{1}
}
\affiliations{
    \textsuperscript{1}University of Central Florida\quad
    \textsuperscript{2}Independent Researcher\\
    zahra.khodagholi@ucf.edu
}

\begin{document}

\maketitle

% ====================================================================
% ABSTRACT
% ====================================================================
\begin{abstract}
Implicit and equilibrium neural layers define their output as the fixed point of a
map. Existing theory certifies that this fixed point exists; it does not certify how
it responds to its inputs. For a conservation-law layer with order-preserving
coupling we give a form-independent answer: every input-output comparative static is
sign-certified if and only if the residual Jacobian is a nonsingular M-matrix, and
the certified sign depends only on the sign of the occupancy derivative, so one
guarantee covers Langmuir, Hill, and saturating-exponential occupancies. The
M-matrix machinery is classical; the contribution is form-independence as the
portable invariant, the first certified-sign sensitivity result for a layer of this
class, and a sharp phase boundary at which certification is lost while the
equilibrium remains well posed. Three consequences follow: the phase boundary, which
is strictly stronger than the M-matrix condition; competitive derepression, whose
derivative is identically zero for any independent per-target scorer and provably
nonzero here; and non-submodularity of resource-coupled suppression. We instantiate
on siRNA off-target competition and validate on real public data only, across
seventeen microRNAs, two titration studies, and a paired chemical-modification
experiment; a second, non-biological instantiation shows the same certified sign on
measured physical data. As a trainable layer the operator learns end-to-end while
remaining certified-monotone by construction, reaching magnitude parity with a
parameter-matched feedforward baseline while expressing a competition effect that
baseline cannot represent for any weights. The single-pool instantiation validates
the mechanism but does not rank off-target magnitude or resolve position-specific
design. Code: \url{https://github.com/shadi97kh/certified-equilibrium-layers}
\end{abstract}

% ====================================================================
% INTRODUCTION
% ====================================================================

\begin{figure*}[t]
\centering
\includegraphics[width=\textwidth,height=0.26\textheight,keepaspectratio]{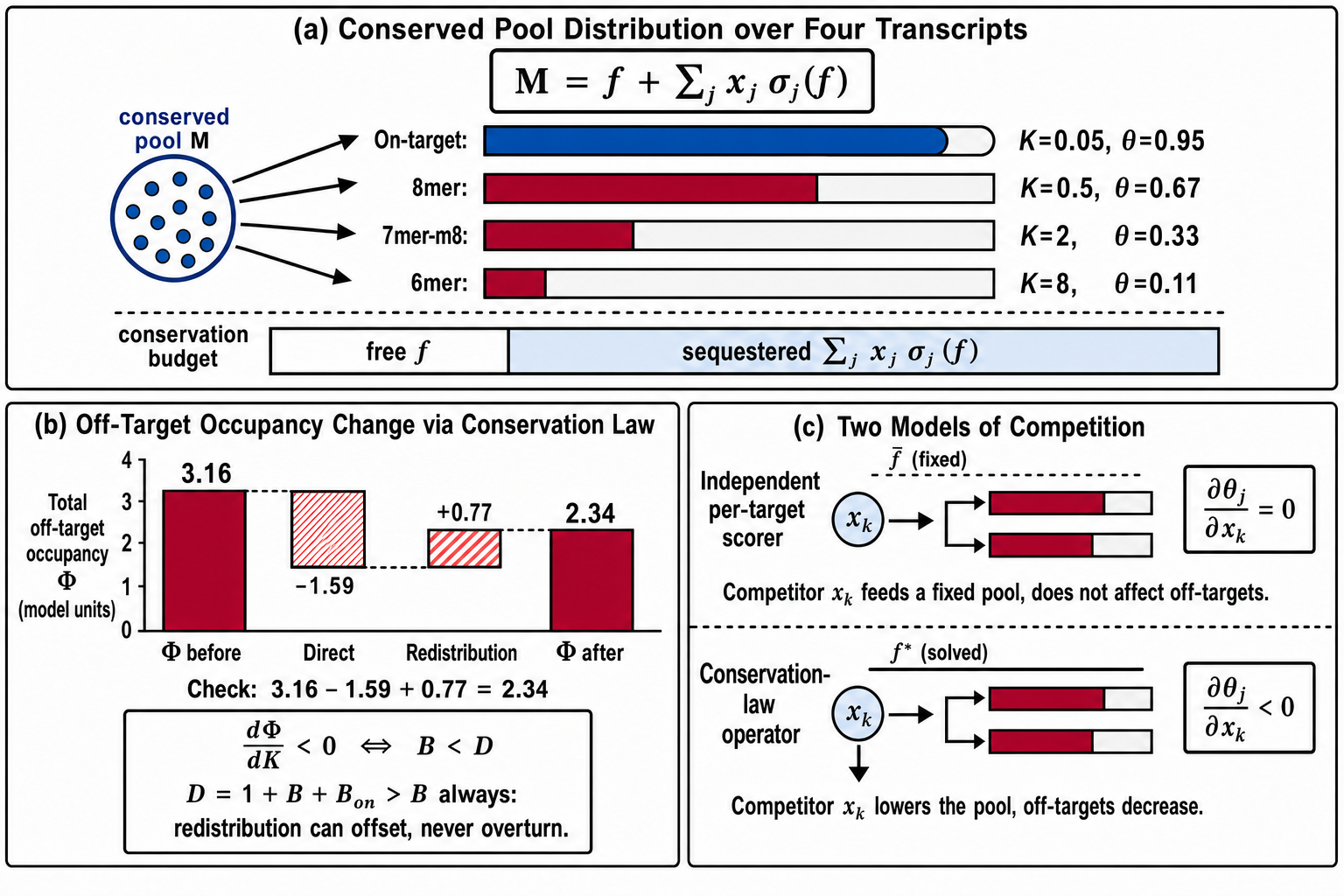}
\caption{The conservation-law equilibrium operator. \textbf{(a)} A conserved pool
is distributed among competing sites; the free amount $f$ solves
$M = f + \sum_j x_j\sigma_j(f)$, so raising one $K$ shifts $f$ and moves every
other occupancy. \textbf{(b)} Raising a seed $K$ lowers $\Phi$ directly ($-1.59$)
while redistribution adds part back ($+0.77$), a net $3.16 \to 2.34$; the certified
condition $B^{\mathrm{off}}<D$ (printed $B$ in the panel) guarantees the offset can
never overturn it. \textbf{(c)} A competitor's abundance is not an argument of an
independent scorer, so its predicted response is identically zero
(Proposition~\ref{prop:derep}); the operator solves for $f^\star$, which falls when
a competitor is added.}
\label{fig:operator}
\end{figure*}

\section{Introduction}

Competition for a conserved, limited resource recurs across the natural and social sciences: a fixed pool is split among many consumers by affinity, and one measures the resulting equilibrium occupancy. Enzyme substrates for a catalyst, ligands for receptors, bidders for divisible supply, and our detailed case, guide RNAs for loaded Argonaute protein, all share this form. The defining feature is coupling: changing one consumer's demand shifts the free resource available to every other, so consumers cannot be treated independently.

Machine learning has a natural object for equilibrium computation, the implicit or deep-equilibrium layer, whose output is the fixed point of a learned map rather than the result of a fixed number of explicit steps \citep{bai2019deq,gu2020ignn}. Well-posedness (existence, uniqueness, and convergence of that fixed point) is well understood for several operator classes \citep{gabor2024pcdeq,yates1995,lemmens2012}. That line answers whether the equilibrium is well defined, not how it responds to its inputs, nor whether that response can be certified.

This paper answers that second question for conservation-law layers. Comparative statics are governed by the fixed-point Jacobian, which is a Z-matrix; within that class inverse-positivity coincides with the M-matrix property. So every comparative static is sign-certified, its sign fixed by the coupling topology independent of affinity and abundance magnitudes, iff the Jacobian is a nonsingular M-matrix. The characterization is form-independent and disjoint from the existence-and-uniqueness results above.

Three consequences follow, each expressing something an independent per-consumer scoring function cannot represent even in principle: a phase boundary at which certified design is lost, entered only under positive loading feedback and lying outside the physically supported RNAi regime, so it is a mathematical characterization the real data never cross; competitive derepression, whose derivative with respect to a competitor's abundance is identically zero for any independent scorer and nonzero for the operator; and the failure of submodularity under coupling, which removes the guarantee that would license greedy design.

We instantiate on RNA interference: the shared resource is the loaded Argonaute pool, the consumers are guide-RNA target sites, the occupancy is predicted repression. RNAi is now an approved therapeutic modality \citep{adams2018}, and guide design must reason jointly about on-target efficacy and seed-mediated off-target repression, the dominant sequence-driven liability \citep{jackson2003,birmingham2006,schlegel2022}. We validate on real public data only, no synthetic data. The M-matrix characterization holds in all seventeen microRNAs of a transfection corpus to numerical precision and is form-independent. The operator reproduces the direction of competitive derepression measured in two independent titration studies, in a regime where independent scoring methods are provably flat. And a certified seed modification reduces measured off-target repression in a paired before-and-after chemical-modification experiment.

The single-pool instantiation has boundaries, delimited throughout, and they bound the instantiation rather than the operator class. The message for implicit-layer practitioners is that monotonicity need not be imposed architecturally or verified after training: under order-preserving coupling it is a certified, form-independent property of the fixed point that survives learning.

\paragraph{Contributions.}
\begin{itemize}
\item We prove a form-independent certified comparative-statics result for conservation-law equilibrium layers: every comparative static is sign-certified if and only if the residual Jacobian is a nonsingular M-matrix, with the sign depending only on $\mathrm{sign}(\sigma')$ (Theorem~\ref{thm:mmatrix}). The tools are classical \citep{berman1994,topkis1998,galichon2025}; the contributions are form-independence as a portable invariant and the first certified-sign sensitivity result for a layer of this class, where existing implicit-layer theory certifies only well-posedness.
\item We derive three consequences independent per-consumer scoring cannot express: a characterized phase boundary for certified design, sharper than the M-matrix condition (Theorem~\ref{thm:phase}, Proposition~\ref{prop:single}); competitive derepression with an identically-zero independent-scorer derivative (Proposition~\ref{prop:derep}); and non-submodularity of resource-coupled suppression by exact enumeration (Proposition~\ref{prop:submod}).
\item We instantiate on RNA interference and validate on real public data: comparative statics match a finite-difference re-solve to relative error near $10^{-6}$ across seventeen microRNAs; the operator reproduces titration behavior where scoring methods are provably flat; and a certified seed modification reduces measured off-target repression. A second, non-biological instantiation confirms the certified sign is a property of the operator class, and we delimit the instantiation's limits honestly.
\item We instantiate the operator as a \emph{trainable} equilibrium layer whose backward pass is the certified comparative static of \eqref{eq:ift}. Trained end-to-end it remains certified-monotone by construction at every step, reaching parity with a parameter-matched feedforward baseline while expressing a competition effect the baseline structurally cannot: certified monotonicity is a property to train under, not to impose architecturally or verify post hoc \citep{you2017lattice,liu2020certmono,wehenkel2019umnn,sivaraman2020}.
\end{itemize}

% ====================================================================
\section{Related Work}

\paragraph{Implicit and equilibrium layers.}
Deep equilibrium models define their output as a fixed point and differentiate through it with the implicit function theorem \citep{bai2019deq}. Implicit graph neural networks extend this to graph-structured propagation \citep{gu2020ignn}, and monotone operator theory guarantees a unique equilibrium \citep{winston2020mon,baker2023mignn}, where ``monotone'' refers to the operator, which secures well-posedness, and not to the input-output map, which is what we certify. The implicit framework of \citet{elghaoui2021implicit} and the nonlinear Perron-Frobenius analysis of \citet{piotrowski2024fixed} likewise establish well-posedness rather than sensitivity structure. Most relevant, positive-concave deep equilibrium models use nonlinear Perron-Frobenius theory and standard interference functions to establish existence and uniqueness for a class of positive, concave maps \citep{gabor2024pcdeq,yates1995,lemmens2012}. That literature certifies that the fixed point exists and is unique; we certify the sign structure of its sensitivities, and claim no novelty for well-posedness.

\paragraph{Certified monotonicity and comparative statics.}
Monotonicity in inputs is usually imposed, either architecturally through constrained-monotone networks and lattice models \citep{you2017lattice,wehenkel2019umnn,runje2023} or by verification after training \citep{liu2020certmono,sivaraman2020}. Here it is derived. The tools are classical: within the Z-matrix class, inverse-positivity coincides with the nonsingular M-matrix property \citep{berman1994,plemmons1977}, inverse isotonicity extends to nonlinear M-functions \citep{more1973}, and the link to monotone comparative statics is established in mathematical economics \citep{topkis1998}, recently unified for equilibrium problems by \citet{galichon2025} via a gross-substitutes condition. We claim none of this machinery as new. The delta is the threefold contribution listed above: form-independence, a certified-sign sensitivity result where the implicit-layer literature certifies only well-posedness \citep{winston2020mon,gabor2024pcdeq}, and a sharp phase boundary.

\paragraph{siRNA efficacy and off-target prediction.}
On-target efficacy is predicted by sequence models \citep{bai2024oligoformer}; seed-mediated off-target repression by seed-matching and thermodynamic tools \citep{garcia2011,cazares2024} and by affinity models from Argonaute binding assays \citep{mcgeary2019,soutschek2022}. All are per-target scoring functions that score each transcript in isolation, so they are structurally unable to express competition effects
(Proposition~\ref{prop:derep}; Appendix~A.5 and Table~2 of the supplement), though target abundance is known to dilute small-RNA activity at the system level \citep{salmena2011,arvey2010}. Appendix~A surveys this class in full.

\begin{table}[t]
\centering
\small
\setlength{\tabcolsep}{4pt}
\begin{tabular}{@{}l@{\hspace{6pt}}p{0.60\columnwidth}@{}}
\toprule
Symbol & Meaning \\
\midrule
$M_i$ & conserved total resource (loaded RISC pool) \\
$f$, $f^\star$ & free resource; its equilibrium value \\
$x_{is}$, $K_{is}$ & abundance, dissociation constant of site $s$ \\
$\sigma_{is}(f)$ & occupancy: nonneg., increasing, bounded \\
$\theta^\star_{is}$ & equilibrium occupancy $\sigma_{is}(f^\star_i)$ \\
$\Phi_i$ & total off-target occupancy,
  $\sum_{s\in\mathcal{S}^{\mathrm{off}}_i} x_{is}\theta^\star_{is}$ \\
$R$, $J$ & conservation residual; its Jacobian $\partial R/\partial f$ \\
$D_i$ & Jacobian diagonal $1+\sum_s x_{is}\sigma'_{is}\ (\geq 1)$ \\
$B^{\mathrm{off}}_i$, $B^{\mathrm{on}}_i$ & off- and on-target sensitivity
  sums; $D_i = 1 + B^{\mathrm{off}}_i + B^{\mathrm{on}}_i$. Two-pool
  statements write $B_0 := B^{\mathrm{off}}_0$ \\
$c'_i$ & loading feedback $dc_i/df_i$ (two-pool only) \\
$r_0$ & direct suppression rate
  $f_0\sum_{s\in\mathrm{mod}} x_{0s}\sigma'_{0s} > 0$ \\
$B_0(J^{-1})_{00}$ & certified-design margin; certified when ${\leq}\,1$ \\
\bottomrule
\end{tabular}
\caption{Notation. Index $i$ ranges over guides (equivalently, resource
pools), $s$ over the target sites of guide $i$, and $k$ over perturbed
competitors; in single-pool statements we drop $i$ and write
$\sigma_j, x_j, K_j$ for a single consumer $j$. Standing assumptions
(A1) to (A3) are in Appendix~B.}
\label{tab:notation}
\end{table}

% ====================================================================
\section{The Conservation-Law Equilibrium Operator}

\paragraph{Setup.}
Let $n$ consumers compete for a conserved resource of total amount $M > 0$. Let $f \ge 0$ be the free (unbound) resource at equilibrium, $x_j \ge 0$ the abundance of consumer $j$, and $\sigma_{j}(f)$ its occupancy function, nonnegative, increasing in $f$, and bounded. Conservation is the constraint
\begin{equation}
M \;=\; f \;+\; \sum_{j} x_j\, \sigma_{j}(f),
\label{eq:conservation}
\end{equation}
total resource on the left, free resource plus what consumers sequester on the right. The canonical Langmuir occupancy is $\sigma_j(f)=f/(K_j+f)$, $K_j$ the dissociation constant of consumer $j$; we also use Hill and saturating-exponential forms. Table~\ref{tab:notation} collects the notation; appendix references throughout are to the supplement.

\paragraph{The operator.}
We define the conservation-law equilibrium operator
\begin{equation}
\mathcal{E}(K, x, M) \;\longmapsto\; (f^\star, \theta^\star),
\label{eq:operator}
\end{equation}
where $f^\star$ solves \eqref{eq:conservation} and $\theta^\star_j = \sigma_j(f^\star)$. Under the assumptions $f^\star$ exists, is unique and interior, and is found by bisection on the scalar residual (Appendix~C); existence and uniqueness for positive-concave layers of this type are standard \citep{lemmens2012,gabor2024pcdeq}. The object of interest is the map from inputs to $(f^\star, \theta^\star)$ and its derivatives.

The multi-pool generalization, in which several conserved resources couple through a shared loading process, replaces the scalar constraint by a system of conservation equations and is where the phase boundary lives; we develop it in Appendix~F and use it in Theorem~\ref{thm:phase}, in contrast to the single-pool case of Proposition~\ref{prop:single}.

% ====================================================================
\section{Certified Comparative Statics}

Write the residual of \eqref{eq:conservation} as $R(f, x) = f + \sum_j x_j \sigma_j(f) - M$, with Jacobian $J = \partial R/\partial f$ in the multi-consumer (vector $f$) formulation of the supplement. By the implicit function theorem,
\begin{equation}
\frac{\partial f}{\partial x_k} \;=\; -\,J^{-1}\,\frac{\partial R}{\partial x_k},
\qquad
\left(\frac{\partial R}{\partial x_k}\right)_i \;=\; \sigma_{k}(f)\,\ge 0 .
\label{eq:ift}
\end{equation}
Every sensitivity is $-J^{-1}$ applied to a nonnegative vector, so certified signs reduce entirely to the sign structure of $J^{-1}$.

Two terms recur. A comparative static is \emph{sign-certified} if its sign is determined for all nonnegative occupancy forcings and all positive abundances, independent of the magnitudes of $K$ and $x$. The coupling is \emph{order-preserving} if increasing any consumer's free-resource share does not increase another's sequestration, equivalently if the off-diagonal entries of $J$ are nonpositive, so that $J$ is a Z-matrix. Both are stated formally in Appendix~E.

\begin{theorem}[Certified comparative statics]
\label{thm:mmatrix}
Consider a conservation-law equilibrium layer with order-preserving coupling, so that the residual Jacobian $J$ is a Z-matrix. The following are equivalent:
\begin{enumerate}
\item[(i)] $J$ is a nonsingular M-matrix;
\item[(ii)] $J^{-1} \ge 0$ entrywise (inverse-positivity);
\item[(iii)] every comparative static $\partial f_i/\partial x_k$ is sign-certified, with $\partial f_i/\partial x_k \le 0$ for all nonnegative forcings.
\end{enumerate}
The equivalence is independent of the occupancy form: it holds identically when each $\sigma_j$ is Langmuir, Hill, or saturating-exponential, and more generally for any nonnegative, increasing, bounded, differentiable $\sigma_j$.
\end{theorem}

\begin{proof}[Proof sketch]
Order-preserving coupling makes the off-diagonal entries of $J$ nonpositive, while the
diagonal $J_{ii}=1+\sum_s x_{is}\sigma'_{is}\ge 1$ is strictly positive, so $J$ is a
Z-matrix. Within that class, inverse-positivity and the nonsingular M-matrix property are
equivalent \citep[Ch.~6, Thm.~2.3]{berman1994}, giving (i)$\Leftrightarrow$(ii). For
(ii)$\Rightarrow$(iii), \eqref{eq:ift} writes every comparative static as $-J^{-1}b$ with
$b\ge 0$, so $J^{-1}\ge 0$ fixes the sign, and the chain rule
$\partial\theta^\star_{is}/\partial x_k = \sigma'_{is}\,\partial f^\star_i/\partial x_k$
transfers it to occupancies because $\sigma'\ge 0$. Conversely, a negative entry
$(J^{-1})_{pq}<0$ admits the admissible forcing $b=e_q$, giving $\partial
f^\star_p/\partial x_q>0$ and contradicting certification. Form-independence follows
because every step uses only $\mathrm{sign}(\sigma')$ and the signs of the off-diagonal
couplings, never their magnitudes or the algebraic form of $\sigma$. Full proof in
Appendix~E.
\end{proof}

The result is an exact characterization, not a sufficient condition, because order-preserving coupling supplies the Z-matrix structure under which inverse-positivity and the M-matrix property coincide. Certification is therefore not imposed but holds wherever the coupling is order-preserving, and fails exactly when it does not. The next section characterizes where.

% ====================================================================
\section{Three Consequences}

\subsection{A phase boundary for certified design}

Consider weakening a consumer's affinity, raising its $K$, to reduce its occupancy. In the single-pool operator this is unconditionally safe for the off-target channel.

\begin{proposition}[Single-pool monotonicity, off-target channel]
\label{prop:single}
In the single-pool operator \eqref{eq:conservation}, raising the dissociation constant on a set of a guide's off-target sites, with the on-target sink unchanged, strictly decreases that guide's total off-target occupancy, with no condition on magnitudes: writing $D = 1 + B^{\mathrm{off}} + B^{\mathrm{on}}$ for the Jacobian diagonal, strict diagonal dominance $D > B^{\mathrm{off}}$ holds because the free-resource term is strictly positive and the on-target sink is nonnegative, so the certified sign of Theorem~\ref{thm:mmatrix} holds without further conditions.
\end{proposition}

The boundary appears in the multi-pool setting, where guides are loaded onto the shared resource through a competitive process. Let the loading competitiveness of a consumer depend on its free amount through a derivative $c'$. The certified sign can be overturned only when this feedback is positive.

\begin{theorem}[Phase boundary for certified design]
\label{thm:phase}
In the two-pool operator with competitive loading, raising a consumer's seed affinity reduces its off-target occupancy if and only if the closed-form dominance condition $B_0(J^{-1})_{00}\le 1$ holds. The condition is satisfied unconditionally in the single pool; in the coupled case it is implied by, but strictly stronger than, the M-matrix property of the residual Jacobian, so certified reduction can be lost through redistribution while the equilibrium is still well posed (the certified-design boundary lies inside the M-matrix region). The intervention can backfire, increasing total off-target occupancy, only under positive loading feedback ($c'>0$) in the physical regime.
\end{theorem}

\begin{proof}[Proof sketch]
Differentiating the coupled residual totally in the modification parameter $\delta$ gives
$d\Phi_0/d\delta = r_0\big(B_0(J^{-1})_{00}-1\big)$ with $r_0>0$. For the $2\times 2$
coupled $J$, $(J^{-1})_{00}=J_{11}/\det J$, so with $\det J>0$ the certified condition
reads $\det J \ge B_0 J_{11}$. Since $B_0 J_{11}>0$, this threshold is interior to
$\{\det J>0\}$: an M-matrix with $0<\det J<B_0 J_{11}$ still backfires, which is why the
M-matrix property is necessary but not sufficient. Full proof in Appendix~F.
\end{proof}

The three thresholds are not identical: certified reduction is lost at $\det J = B_0 J_{11}$, inverse-positivity at $\det J = 0$, and the interior equilibrium at the same singularity. The certified-design boundary is tightest and is the operative one for design. For RNA interference specifically, the biophysics of Argonaute loading, an ATP-dependent chaperone-mediated assembly step \citep{iwasaki2010}, argues that a matured guide does not re-enter loading competition, placing the physiological regime on the certified ($c'\le 0$) side; we therefore treat the overturned regime as a characterized mathematical boundary rather than an empirical prediction, and validate the certified side.

\subsection{Competitive derepression}

\begin{proposition}[Structural blindness of independent scoring]
\label{prop:derep}
Let an independent per-consumer scorer predict consumer $j$'s occupancy as $g(\text{sites of } j,\, \bar f)$ for a fixed resource level $\bar f$ that does not depend on the abundance $x_k$ of a competitor $k \ne j$. Then $\partial g/\partial x_k = 0$ identically. In contrast, the conservation-law operator satisfies $\partial \theta_j^\star/\partial x_k < 0$: increasing a competitor's abundance lowers the shared free resource and derepresses consumer $j$.
\end{proposition}

Competitive derepression is not a quantity on which the operator and a scorer disagree numerically; a scorer cannot represent it at all, because the competitor's abundance is not in the domain of its scoring function. The derivative is exactly zero by the signature of the function, for any calibration. This is quantitative, not only structural: the gap no calibration removes is bounded below by half the induced occupancy range (Appendix~E), which is $0.342$ on the miR-122 conditions as the operator predicts them. The operator over-predicts the upper endpoint ($0.71$ against a measured $0.5$), so the realized bound there is nearer $0.24$; the structural claim, that the bound is strictly positive for every calibration, does not depend on either value.

\subsection{Non-submodularity under coupling}

\begin{proposition}[Non-submodularity]
\label{prop:submod}
Let the design objective be total off-target suppression as a function of the set $S$ of affinity-weakening interventions. In the independent-site model this set function is submodular in the physiological (subsaturated) regime, where each site's suppression curve is concave. Under the shared-resource coupling of \eqref{eq:conservation} it is not submodular in any loading regime we enumerate: resource redistribution induces increasing marginal returns for some interventions. Consequently the greedy $1-1/e$ guarantee \citep{nemhauser1978} does not apply, and on the small (seed-sized) design space exact combinatorial optimization returns the true optimum.
\end{proposition}

We establish this by exact enumeration over the full lattice (all $2^{6}$ subsets and all diminishing-returns triples) on the coupled equilibrium, contrasting it with the independent-site model where submodularity holds in the subsaturated regime; the regime table is in Appendix~G.

% ====================================================================
\section{Instantiation on RNA Interference}

We instantiate $\mathcal{E}$ with the loaded Argonaute-RISC pool as $M$, guide-RNA target sites as consumers, per-site dissociation constants $K$ from Argonaute binding data, and equilibrium occupancy as predicted repression. Off-target repression is occupancy on transcripts whose $3'$UTRs carry a guide seed match; a thermally destabilizing seed modification raises the corresponding $K$. This is the setting for all experiments below, where both the strengths and the boundaries of the single-pool model appear. The affinity source, the seed-matching, the multi-pool loading model, and the two algorithms (equilibrium solve and certified design) are in Appendices~C, D, and F; the certificate in the design procedure is automatic in the single pool (Proposition~\ref{prop:single}) and binding only under two-pool coupling.

A seed modification in practice also weakens the fully paired on-target site.
Proposition~\ref{prop:single} covers the off-target channel; when the on-target sink is
weakened enough to release guide, the released guide can raise total off-target occupancy
in the guide-limited regime (Appendix~G). We report this rather than absorb it into the statement of Proposition~\ref{prop:single}.

\paragraph{What the certificate changes for a designer.}
Three consequences are practical rather than theoretical. The certified direction is calibration-free across a sweep of the affinity ratio over $[1.1,10]$, so a destabilizing chemistry can be committed to before its thermodynamics are measured. The antecedent acts as a screen: chemistries that provably raise seed $K$ are certified; those perturbing Argonaute-channel geometry rather than base pairing are not, and there the operator makes no prediction. And off-target burden depends on the abundance vector, not seed-match counts alone, so the same guide carries different burden in different expression contexts, a dependence no independent scorer can express for any calibration (Proposition~\ref{prop:derep}). None of these requires the magnitude ranking the single-pool model lacks.

% ====================================================================
\begin{figure*}[t]
\centering
\includegraphics[width=0.86\textwidth]{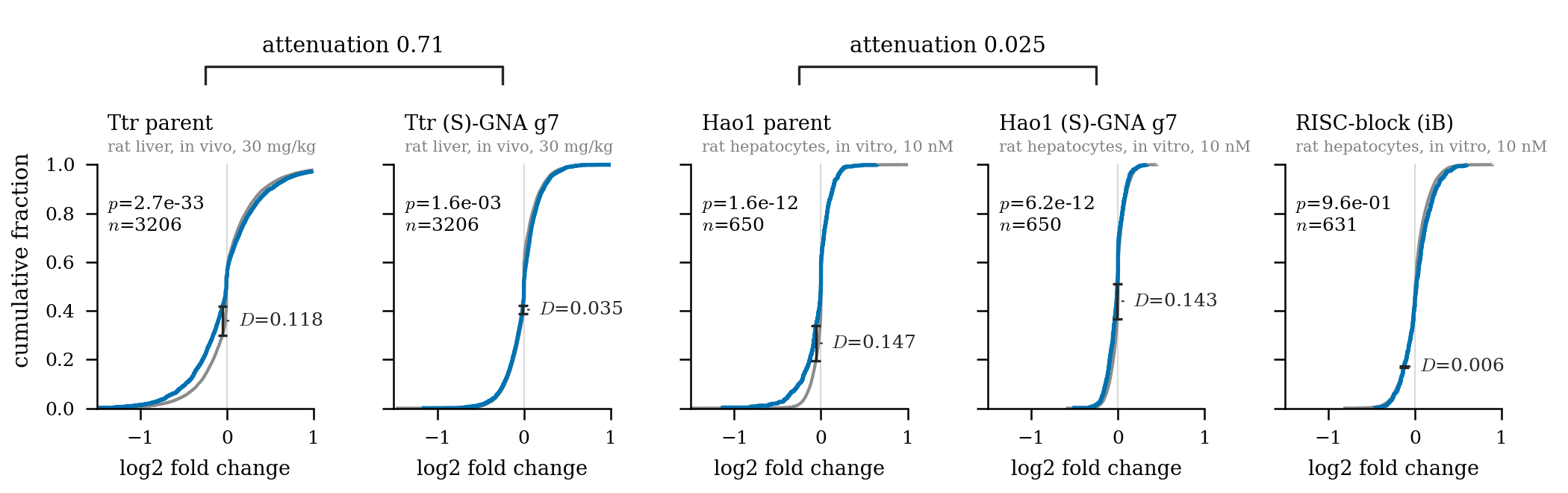}
\caption{Certified-direction panel. Empirical CDF of off-target log2 fold
change for seed-matched genes (blue) against non-matched background (grey),
with identical matched and background sets across both arms of each contrast;
$D$ is the vertical separation at maximum gap. Left: the parent Ttr guide
represses seed-matched genes strongly and a single (S)-GNA substitution at
position seven largely abolishes the effect. Centre: the same chemistry on a
second guide barely moves the signature; assay format and dose differ as well
as guide, so we do not attribute the difference to the guide alone. Right:
blocking RISC loading abolishes the signature, the mechanistic control for the
conserved loaded-pool premise. The blocked arm is from a different sequencing batch than the Hao1 panels;
its same-batch parent is reported in the text. Axis clipped to $[-1.5,1.0]$ for display; $D$,
$p$, and attenuation computed unclipped from unrounded $D$.}
\label{fig:certified}
\end{figure*}

\section{Experiments}

All experiments use real public data; loaders raise on missing inputs rather than substitute, and no synthetic data is used at any stage. We report the following studies, mapped to the results above. Full protocols and accessions are in Appendix~K.

\paragraph{Theory verification at scale.}
We verify Theorem~\ref{thm:mmatrix} on the microRNA transfection corpus of \citet{mcgeary2019} (accession GSE140217), forming the residual Jacobian at the real equilibrium for each of the seventeen microRNAs from measured $3'$UTR seed complementarity and per-site affinities. Single-microRNA transfection probes the single-pool regime, where the Jacobian is diagonal and positive, hence a nonsingular M-matrix by construction (Proposition~\ref{prop:single}); we therefore do not treat ``M-matrix in $17/17$'' as a test, since it cannot fail. The sign-flip boundary lives in the multi-pool regime and is characterized separately in Theorem~\ref{thm:phase}. The non-vacuous checks are numerical: the closed-form comparative static matches a forward finite-difference re-solve to maximum relative error near $10^{-6}$, and the certified sign is unchanged under Hill and saturating-exponential occupancies, confirming form-independence on real affinities. The agreement is looser for the Hill form (near $10^{-4}$) than the two concave forms, tracking occupancy concavity, not the sign, which is form-independent throughout. A cross-check using per-site dissociation constants from an independent affinity model \citep{soutschek2022} reproduces the result, so it is not an artifact of the affinity source.

\paragraph{A capability independent scoring cannot express.}
We set the abundance regime from \citet{denzler2014} and \citet{bosson2014} and compute the onset thresholds with no further fitting; the miR-122 threshold falls within a factor of $1.6$ of the measured anchor. That anchor is a separate measurement from the same study that supplies the miR-122 stoichiometry, so this is a consistency check and we do not present it as a held-out prediction. Sweeping competitor abundance for a focal target, the independent per-target scorer predicts exactly zero derepression across the entire sweep in both scenarios, because competitor abundance is not one of its arguments; the derivative is identically zero and this is calibration-free. The operator predicts derepression rising with competitor abundance and, at each study's measured regime, agrees in direction with the observed release. The two measured thresholds differ about fiftyfold ($1.5\times10^{5}$ versus $3\times10^{3}$); the operator orders them correctly but compresses the separation to about fivefold ($2.4\times10^{5}$ versus $4.5\times10^{4}$; Appendix~K), over-predicting the miR-92/25 threshold fifteenfold. We claim the ordering, not the separation. The absolute abundances are a stoichiometric input, not a fit, so the exact threshold is calibration-dependent; the load-bearing, calibration-free part is the structural flatness of the independent scorer, which cannot express the effect (operator log-slope $0.874$ for miR-122 and $0.674$ for miR-92/25, against $0.000$ for every independent scorer; Appendix~K). The ordering is robust to that calibration: over three orders of magnitude of affinity scale the operator ranks miR-122 above miR-92/25 at every scale, the separation compressing from about $18\times$ to $3\times$. That sweep uses a derepression-crossing threshold rather than the tabulated one, so it certifies the ordering and not the tabulated magnitudes (Appendix~K).

\paragraph{Certified design on paired chemical-modification data.}
For each parent/modified pair we annotate $3'$UTRs for canonical $6$-$8$mer seed matches, compute the one-sided Kolmogorov-Smirnov statistic $D$ for a left shift of seed-matched genes against background \emph{within each arm}, and set attenuation to $1 - D_{\mathrm{mod}}/D_{\mathrm{parent}}$. Two invariants are enforced in code: both arms use the same control, and both are scored on one gene universe shared by the pair, with identical matched and background sets. A direct modified-versus-parent contrast is invalid: a global potency difference propagates into every gene's fold change and manufactures an apparent seed effect, which the arm-versus-control design cancels.

On the Schlegel rat-liver GalNAc-siRNA data \citep{schlegel2022} (GSE184929), a therapeutic guide targeting \emph{Ttr} carrying a single (S)-GNA seed modification at guide position seven is compared to its parent at matched $30$ mg/kg dose against a shared vehicle group. The parent shows a strong seed-mediated left shift ($D = 0.118$, $p = 2.7\times10^{-33}$); after modification the shift is largely abolished ($D = 0.035$, $p = 1.6\times10^{-3}$), an attenuation of $0.71$ (bootstrap $95\%$ CI $[0.57, 0.86]$) on $3{,}206$ seed-matched genes. The certified direction holds and is stable to the $3'$UTR isoform annotation, the site-class definition, and the seed annotator. As an external reproduction, SeedMatchR \citep{cazares2024} on its $7$mer-m8 class gives parent $D = 0.107$ and modified $D = 0.034$ on the same data; our annotator sits between SeedMatchR's own two site definitions, so the site-definition choice moves $D$ by more than the annotator implementation does.

\paragraph{Contrasts that cannot answer the question, and the scope of the certificate.}
We report two further contrasts that are unmeasurable rather than negative: one confounded with batch, one whose seed matches $46\%$ of human $3'$UTR genes. Neither supports or refutes the certified direction (Appendix~K.7). The certificate also requires a map from chemistry to a change in seed affinity. For (S)-GNA the certified sign is invariant to its exact value, holding throughout a sweep of $K_{\mathrm{mod}}/K_{\mathrm{parent}}$ over $[1.1,10]$, so no calibration carries the claim. For alkyl-phosphonate backbone modifications no measured thermodynamic value establishes an increase in seed $K$, so the certificate's antecedent is not established and the operator makes no prediction there. We adopted this restriction after those contrasts returned no usable signal, and do not present it as pre-specified (Appendix~K.8).

\paragraph{Mechanistic control and per-guide variation.}
Blocking RISC loading abolishes the seed signature ($D=0.006$, $p=0.96$), against $D=0.147$ and $0.180$ for the loaded parent in the study's two sequencing batches, which have different gene universes; both comparisons confirm the conserved loaded-pool premise. On a second guide (Hao1) the same chemistry at the same seed position gives attenuation $0.025$, effectively none, against $0.71$ on Ttr; the certified sign is technically satisfied but the magnitude is indistinguishable from zero, and the verdict depends on the $3'$UTR annotation ($0.025$ under the union-over-isoforms annotation we adopt, $0.298$ under single-isoform). The single-pool model, seeing only aggregate occupancy, does not distinguish the two cases. This is the sharpest statement of the instantiation's magnitude-ranking limitation (Appendix~K.9).

\paragraph{Competitive adsorption: the coupled regime on real data.}
The RNAi instantiation above is single-pool, so its Jacobian is diagonal and the M-matrix property holds trivially. The multi-pool regime, where the M-matrix property is non-trivial and off-diagonal coupling is present, is exercised on competitive Langmuir adsorption of arsenate and fluoride on activated carbon \citep{acharya2024}, a genuinely two-species system with both cross-directions measured. There the residual Jacobian is a Z-matrix and a nonsingular M-matrix at every measured concentration, and the certified sign matches the measured antagonism at every point (sign-match $4/4$): the coupled certification on real data. The magnitude is over-predicted by the single-constant extended Langmuir model; the sign is what the theorem certifies. This is also the one check that could have come out otherwise: multi-component adsorption is sometimes synergistic, a competitor raising rather than lowering another species' loading, which violates order-preserving coupling and voids the theorem's antecedent. The agreement tests that antecedent rather than restating the conclusion. Appendix~I gives the instantiation, the proof, and the figure.

\begin{figure}[t]
\centering
\includegraphics[width=\columnwidth]{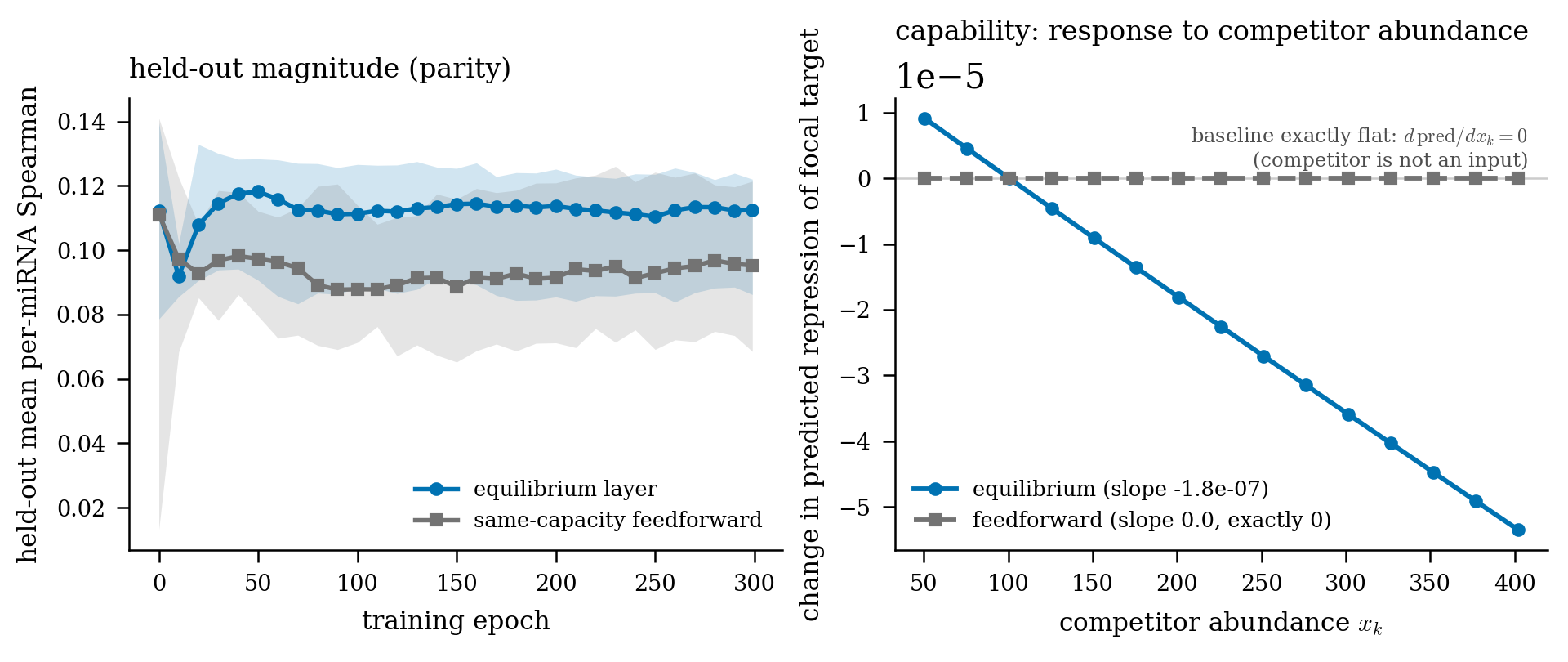}
\caption{Trainable certified-monotone layer on the McGeary corpus (GSE140217),
four held-out microRNAs. \textbf{Left:} held-out magnitude ranking. The equilibrium layer reaches parity with a parameter-matched feedforward baseline (mean per-microRNA Spearman $0.112 \pm 0.013$ vs.\ $0.095 \pm 0.015$ over ten seeds; $4739$ vs.\ $4737$ parameters). The per-seed ranges overlap, so we report parity, not improvement. \textbf{Right:} sweeping a competitor's abundance
$x_k$ for a held-out focal target, the feedforward prediction is bit-identical
(spread exactly $0$: $\partial/\partial x_k \equiv 0$,
Proposition~\ref{prop:derep}), while the equilibrium layer's focal repression
falls (slope $-1.8\times10^{-7}$ per unit $x_k$, the derepression direction).}
\label{fig:learned}
\end{figure}

\paragraph{A trainable certified-monotone layer.}
The comparative static of \eqref{eq:ift} is the implicit-function-theorem gradient of a fixed-point layer, precisely the object a deep-equilibrium model backpropagates through. This lets us make the operator a \emph{trainable} layer and ask whether the certificate survives learning. We replace fixed per-site affinities with a learned map $K_\theta(\cdot)$, a shallow network (one hidden layer, width $64$) on seven site features already in the pipeline, with a softplus output and small floor so $K_\theta>0$ and (A1) holds by construction. The forward pass is the existing bisection solve; the backward pass is the closed-form gradient of \eqref{eq:ift}, a custom differentiation rule rather than differentiation through the solver iterations. We train end-to-end on measured repression in the McGeary corpus \citep{mcgeary2019} (GSE140217), holding out four entire microRNAs so no held-out pool enters training.

Three results follow. First, the certificate holds throughout learning by construction: invariant to the parameters, hence to any training trajectory (Appendix~E). Logged at all $300$ steps, its minimum is $101.18$, far from the boundary. The trained layer is certified-monotone at every step with no architectural constraint and no post-hoc check, the guarantee constrained-monotone networks obtain only by restricting architecture \citep{you2017lattice,runje2023} or verifying after training \citep{liu2020certmono,sivaraman2020}. Second, on held-out magnitude ranking the layer reaches parity with a same-capacity feedforward predictor. Across ten seeds the equilibrium layer averages mean per-microRNA Spearman $0.112 \pm 0.013$ against the baseline's $0.095 \pm 0.015$ (single-seed values $0.118$ and $0.097$, Figure~\ref{fig:learned}, left); both correlations are weak and the per-seed ranges overlap, so we report parity, not improvement. The point is that the certificate and the capability below cost nothing in accuracy. Third, the capability separating the two classes needs no training: sweeping a competitor's abundance for a held-out focal target, the baseline's prediction is bit-identical (derivative identically zero, as Proposition~\ref{prop:derep} requires), while the equilibrium layer's focal repression falls monotonically (Figure~\ref{fig:learned}, right). Training establishes only that expressing this costs no accuracy.

\paragraph{A decision an independent scorer cannot inform.}
Two conditions for the same miR-122 guide differ only in competitor abundance: $3\times10^{4}$ sites per cell (A) and the measured IC50 at $4.5\times10^{5}$ (B), the half-maximal point, distinct from the $1.5\times10^{5}$ onset threshold of the earlier study; both real anchors from \citet{denzler2016}. Every independent scorer we tested returns a bit-identical prediction for A and B. The operator predicts $0.027$ at A and $0.711$ at B, an ordering the measured titration confirms. The operator's derepression is monotone in competitor abundance by the M-matrix property, so it necessarily orders B above A and this is not a falsification test; what it shows is that the ordering is one no independent scorer can express for any calibration (Appendix~K.10).

% ====================================================================
\section{Discussion and Conclusion}

We introduced the conservation-law equilibrium layer and proved its input-output
sensitivities are certified, within the order-preserving regime, exactly when the residual
Jacobian is an M-matrix, a characterization independent of the occupancy form and disjoint
from existing well-posedness theory. The most decision-relevant of the three consequences we derive, competitive derepression, is a capability difference provable from the signature of a scoring function
rather than a contest of accuracy. Instantiated on RNA interference and validated on real
public data, the operator reproduces measured competition biology where scoring methods are
structurally flat and supports a certified design intervention. Beyond it: a phase boundary strictly interior to the well-posedness region, and a certificate that survives end-to-end training by construction rather than verification.

The instantiation's boundaries guide future work. The single-pool model does not rank off-target magnitude against seed-based baselines and is position-blind for seed design, because aggregate occupancy under canonical pairing is nearly invariant to which seed nucleotide is weakened. Resolving this needs the position-specific guide geometry in the Argonaute channel that the aggregate model omits. Since certification concerns the operator class rather than the Langmuir form, any structurally resolved occupancy that stays order-preserving inherits it; supplying one is a concrete next step. The
multi-pool regime, where the M-matrix property is non-trivial and off-diagonal coupling is present, is exercised here on the competitive-adsorption instantiation, whose two-species
coupling is measured; the RNA-interference instantiation is single-pool, and testing the coupled operator on RNAi awaits multi-guide competition data resolving per-guide occupancy in a shared pool.

% References

\bibliography{references}

\setcounter{theorem}{2}
\setcounter{proposition}{3}
\setcounter{corollary}{0}
\setcounter{lemma}{0}
\setcounter{definition}{0}
\setcounter{remark}{0}
\setcounter{figure}{3}
\setcounter{table}{1}
\setcounter{equation}{3}

\appendix
\onecolumn
\setcounter{secnumdepth}{2}
\setcounter{tocdepth}{2}

\section*{Supplementary Material}
\noindent
This supplement contains the full proofs of the results stated in the main paper, the
biological background, the algorithms, the experimental protocols, and the
reproducibility record. References to ``Theorem 1'', ``Theorem 2'', and
``Propositions 1 to 3'' point to the main paper. The main paper is self-contained;
this material is provided for completeness.

\section{Background: siRNA, miRNA, and Seed-Mediated Off-Target Effects}
\label{app:biology}

This section gives the minimal biological background needed to read the rest of the paper. It is written for a machine-learning audience and introduces only the concepts our operator formalizes.

\subsection{Guides, targets, and the silencing complex}

Small interfering RNA (siRNA) and microRNA (miRNA) are short RNA molecules, roughly $21$ to $23$ nucleotides long, that direct sequence-specific repression of messenger RNA (mRNA). Both act through the same molecular machine. One strand of the small RNA, the \emph{guide}, is loaded into an Argonaute (AGO) protein to form the RNA-induced silencing complex (RISC). The loaded guide then base-pairs with complementary sites in target mRNAs. When pairing is extensive, the target is cleaved; when pairing is confined to a short region, the target is destabilized and translationally repressed without cleavage. In therapeutic RNAi, an siRNA is designed to silence one intended (on-target) transcript; endogenous miRNAs instead regulate large networks of transcripts as part of normal gene regulation.

\begin{figure*}[t]
\centering
\includegraphics[width=\textwidth]{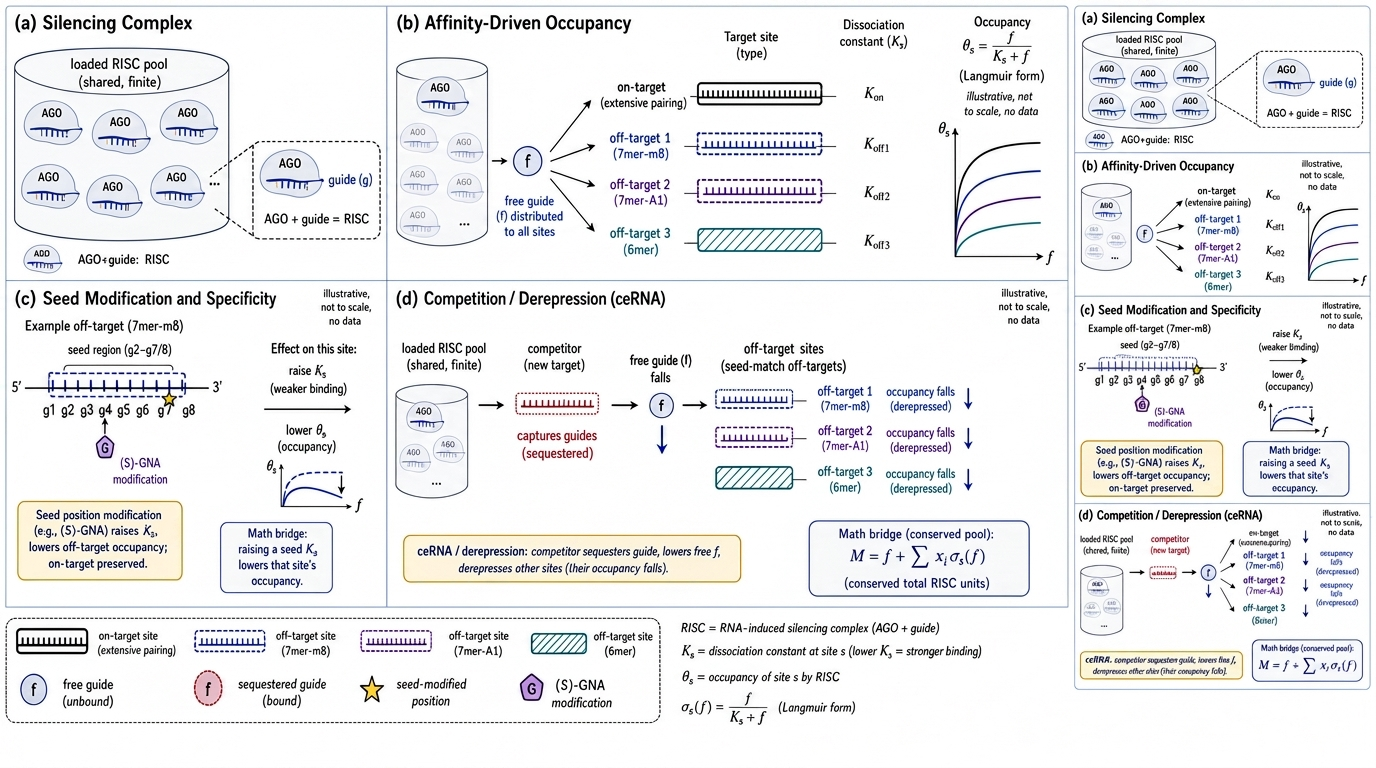}
\caption{Schematic of the biology the operator formalizes (illustrative,
no data). (a) A guide strand loaded into Argonaute forms RISC; the loaded
pool is shared and finite. (b) The same free guide $f$ is distributed over
one extensively paired on-target and several seed-match off-targets
(8mer, 7mer-m8, 7mer-A1, 6mer), each with dissociation constant $K_s$ and
Langmuir occupancy $\theta_s = f/(K_s+f)$. (c) A single (S)-GNA seed
modification raises $K_s$ and lowers that site's occupancy, on-target
preserved. (d) Adding a competitor sequesters guide, lowers free $f$, and
derepresses the other sites (their occupancy falls), the ceRNA effect,
governed by the conserved pool $M = f + \sum_j x_j\sigma_j(f)$. Curves are
schematic; the operator and its results are in the main text.}
\label{fig:background}
\end{figure*}

Two features of this machine are central to our model. First, the guide is a \emph{shared, limited resource}: a finite pool of loaded AGO-guide complex is distributed across all of the sites it can bind, so those sites compete. Second, binding is governed by \emph{affinity}: each potential site $s$ has a dissociation constant $K_s$, and the fraction of time the site is occupied is an increasing, saturating function of the free (unbound) guide concentration $f$, well described by a Langmuir isotherm $\theta_s = f/(K_s+f)$. These two facts, competition for a conserved pool and affinity-driven occupancy, are exactly what the conservation-law operator of the main text encodes; Figure~\ref{fig:background} sketches both.

\subsection{The seed region and off-target effects}

Repression strength is dominated by a short stretch of the guide called the \emph{seed}, nucleotides $2$ through $7$ or $8$ counting from the guide's $5'$ end. A target carrying a contiguous Watson-Crick match to the seed in its $3'$ untranslated region (UTR) is a \emph{seed site}. Seed sites are graded by type: an $8$mer (match to guide positions $2$ through $8$ plus an anchoring adenosine) is strongest, followed by $7$mer-m8, $7$mer-A1, and the weakest $6$mer. Because a match of only six to eight nucleotides occurs frequently by chance across the transcriptome, a guide designed against one intended transcript will also carry seed complementarity to many unintended transcripts.

This is the \emph{off-target} problem. Seed-mediated off-target repression is the principal specificity liability of therapeutic siRNA: an siRNA silences not only its intended target but a background of transcripts that happen to contain seed sites for its guide. Off-target repression per transcript is typically modest (often less than two-fold), but it is broad, affecting hundreds of transcripts, and it is a major source of toxicity in RNAi drugs. Predicting and minimizing it is a central problem in siRNA design, and it is the biological problem our certified-design results address.

\subsection{Seed-modification chemistry}

Because off-target binding is driven by seed affinity, a natural mitigation is to \emph{weaken} the seed: chemically modify one seed-position nucleotide so that seed pairing is destabilized (the dissociation constant $K$ is raised), reducing occupancy at off-target sites. Several chemistries achieve this, including $2'$-O-methyl substitution, glycol nucleic acid (GNA), and other thermally destabilizing modifications. A well-documented example, which we use for validation, is a single (S)-GNA substitution at guide position $7$, which measurably lowers off-target repression while preserving on-target potency. Our certified-design theory formalizes exactly this principle: raising a seed dissociation constant is a certified off-target reducer within the operator's monotone regime. We validate the direction of this effect against real before-and-after RNA-seq measurements.

A subtlety worth noting for interpreting our design results: the effect of a seed modification is \emph{position-specific} in a way that pure thermodynamics does not fully explain. Modifications at adjacent seed positions can produce similar changes in binding stability yet very different off-target outcomes, because the geometry of the guide within AGO makes some positions more consequential than others. A model built on canonical seed pairing alone, including ours in its single-pool form, does not resolve this position specificity; we state this limitation explicitly where it bears on our design validation.

\subsection{Competition and derepression (the ceRNA effect)}

Because loaded guide is a shared limited resource, the targets of a given miRNA are coupled: they compete for the same pool. A direct consequence is \emph{competitive derepression}, also called the competing-endogenous-RNA (ceRNA) effect. If the abundance of one target (or a competing transcript) increases, it sequesters more of the guide, lowering the free guide available to other targets, which are consequently \emph{derepressed} (their repression is relieved). This effect is a pure redistribution phenomenon: it does not change the sequence or sites of the derepressed transcript at all, only the shared resource it competes for.

Whether this effect is physiologically significant was assessed quantitatively in two influential studies. One measured that relieving repression of a highly expressed miRNA (miR-122) required adding on the order of $1.5\times10^5$ competing target sites per
cell \citep{denzler2014}, a high threshold reflecting an abundant miRNA with
large spare capacity. The other found that a miRNA family present at low
abundance relative to its targets (miR-92/25) could be affected by roughly
$3{,}000$ added high-affinity sites \citep{bosson2014}, a much lower threshold. These two measured thresholds, about fiftyfold apart and set by the miRNA-to-target stoichiometry, are the biological anchor of our comparative-statics validation. They also illustrate why competitive derepression cannot be captured by methods that score each target independently: an independent per-target predictor does not take competitor abundance as an input, so its predicted repression of an unchanged transcript is constant, whereas the measured biology, and our operator, show it changing.

\subsection{What existing off-target predictors do}

Standard siRNA and miRNA off-target predictors are scoring functions on a (guide, transcript) pair. Representative families include seed-site counting weighted by site type; seed-pairing-stability scores based on the thermodynamic stability of the seed duplex; context-feature regressions such as TargetScan; and affinity-based occupancy models such as scanMiR, which compress measured Argonaute binding affinities into a per-site occupancy. These methods differ in their features and in how they are fit, but they share a structural property: each scores a single target in isolation, holding an effective guide level fixed. They therefore cannot express effects that arise from competition for the shared resource, such as competitive derepression, and they cannot make certified statements about how a target's repression responds to a perturbation, because they carry no comparative-statics structure. Our operator is not a more accurate scorer on their own axis; it is a different object that expresses the coupled, competition-dependent effects they cannot, and that carries provable sign guarantees on its input-output sensitivities. Where raw per-transcript magnitude prediction is concerned, we report honestly that our single-pool occupancy score does not beat these simple baselines; our contribution is the certified, competition-aware structure, not accuracy on the independent-scoring task. Table~\ref{tab:positioning-supp} summarizes the comparison.

\begin{table}[t]
\centering
\setlength{\tabcolsep}{4pt}
\begin{tabular}{@{}p{3.3cm}ccc@{}}
\toprule
Method & \shortstack{Comp.\\abund.} & \shortstack{Expr.\\derep.} & \shortstack{Cert.\\sign} \\
\midrule
Seed-count \citep{cazares2024}            & no  & no  & no  \\
Seed-pairing stab.\ \citep{garcia2011}    & no  & no  & no  \\
Context regr.\ \citep{agarwal2015}        & no  & no  & no  \\
scanMiR occ.\ \citep{soutschek2022}       & no  & no  & no  \\
Operator (ours)                           & yes & yes & yes \\
\bottomrule
\end{tabular}
\caption{Where the conservation-law operator differs from the standard
per-target scoring class. Independent scorers score each transcript in
isolation, so competitor abundance is not an input, they cannot express
derepression, and they carry no comparative-statics guarantee. The
operator supplies all three; it does not improve on them at raw
off-target magnitude ranking.}
\label{tab:positioning-supp}
\end{table}

\section{Notation and Standing Assumptions}
\label{app:notation}

We collect the objects used throughout. Let $n$ be the number of guide species (loaded miRNA or siRNA guides sharing a resource pool). For guide $i$, let $\mathcal{S}_i$ index its target sites, with site $s \in \mathcal{S}_i$ carrying dissociation constant $K_{is} > 0$ and target abundance $x_{is} \ge 0$. Let $M_i > 0$ be the conserved total of loaded guide $i$. The free (target-unbound) concentration of guide $i$ is $f_i \ge 0$, collected in $f = (f_1,\dots,f_n)^\top$.

The single-species occupancy of site $s$ by guide $i$ is written $\sigma_{is}(f_i)$. Unless stated otherwise (Appendix~\ref{app:form}), $\sigma$ is the Langmuir form
\begin{equation}
\sigma_{is}(f_i) \;=\; \frac{f_i}{K_{is} + f_i}.
\label{eq:app-langmuir}
\end{equation}
We write $\theta_{is} := \sigma_{is}(f_i)$ for the equilibrium occupancy and $\Phi_i := \sum_{s \in \mathcal{S}_i^{\mathrm{off}}} x_{is}\,\theta_{is}$ for total off-target occupancy of guide $i$ over its off-target site set $\mathcal{S}_i^{\mathrm{off}} \subseteq \mathcal{S}_i$. Table~\ref{tab:notation-full} collects these and the remaining symbols.

\begin{table}[t]
\centering
\small
\setlength{\tabcolsep}{4pt}
\begin{tabular}{@{}l@{\hspace{6pt}}p{0.62\textwidth}@{}}
\toprule
Symbol & Meaning \\
\midrule
\multicolumn{2}{@{}l}{\textit{Main-paper symbols}}\\
$M_i$ & conserved total resource (loaded RISC pool) \\
$f$, $f^\star$ & free resource; its equilibrium value \\
$x_{is}$, $K_{is}$ & abundance, dissociation constant of site $s$ \\
$\sigma_{is}(f)$ & occupancy: nonneg., increasing, bounded \\
$\theta^\star_{is}$ & equilibrium occupancy $\sigma_{is}(f^\star_i)$ \\
$\Phi_i$ & total off-target occupancy,
  $\sum_{s\in\mathcal{S}^{\mathrm{off}}_i} x_{is}\theta^\star_{is}$ \\
$R$, $J$ & conservation residual; its Jacobian $\partial R/\partial f$ \\
$D_i$ & Jacobian diagonal $1+\sum_s x_{is}\sigma'_{is}\ (\geq 1)$ \\
$B^{\mathrm{off}}_i$, $B^{\mathrm{on}}_i$ & off- and on-target sensitivity
  sums; $D_i = 1 + B^{\mathrm{off}}_i + B^{\mathrm{on}}_i$. Two-pool
  statements write $B_0 := B^{\mathrm{off}}_0$ \\
$c'_i$ & loading feedback $dc_i/df_i$ (two-pool only) \\
$r_0$ & direct suppression rate $f_0 A_S > 0$ \\
$B_0(J^{-1})_{00}$ & certified-design margin; certified when ${\leq}\,1$ \\
\addlinespace
\multicolumn{2}{@{}l}{\textit{Additional symbols}}\\
$\mathcal{S}_i$, $\mathcal{S}^{\mathrm{off}}_i$ & site index set of guide $i$;
  its off-target subset \\
$A_S$ & modified-site sensitivity sum
  $\sum_{s\in\mathrm{mod}} x_{0s}\sigma'_{0s}$ \\
$A^{\mathrm{off}}_S$, $A^{\mathrm{all}}_S$ & the same sum over modified
  off-target sites, and over all modified sites
  (Remark~\ref{app-rem:modset}) \\
$P$, $p \prec s$ & modifiable seed positions; incidence of
  position $p$ on site $s$ \\
$\delta_p$ & log-affinity increment at position $p$ \\
$g(S)$, $g_{\mathrm{iso}}$ & suppression set function under coupling;
  its isolated-model surrogate \\
$\Delta_j(X)$ & total variation of $\theta^\star_j$ over a competitor
  sweep $[0,X]$ \\
$A$, $\kappa$, $c_i(f_i)$, $C$ & two-pool loading: shared total, baseline,
  competitiveness of guide $i$, shared denominator \\
$q_{\max,j}$, $C_j$, $W$ & Langmuir capacity, aqueous concentration of
  adsorbate $j$, and the adsorption normalizer
  (Appendix~\ref{app:adsorption}) \\
\bottomrule
\end{tabular}
\caption{Notation. The first block collects the symbols of Table~1 of the
main paper, reproduced here for convenience, with $r_0$ written using $A_S$
from the block below; the second block collects the further symbols this
supplement uses. Index $i$ ranges over guides (equivalently, resource
pools), $s$ over the target sites of guide $i$, and $k$ over perturbed
competitors; in single-pool statements we drop $i$. Standing assumptions
(A1) to (A3) follow.}
\label{tab:notation-full}
\end{table}

\paragraph{Standing assumptions.}
\begin{enumerate}
\item[(A1)] \emph{Positivity.} $K_{is} > 0$, $x_{is} \ge 0$, $M_i > 0$ for all $i,s$.
\item[(A2)] \emph{Monotone bounded occupancy.} Each $\sigma_{is}:\mathbb{R}_{\ge 0}\to[0,1)$ is $C^1$, strictly increasing, with $\sigma_{is}(0)=0$ and $\sigma_{is}(f)\to 1$ as $f\to\infty$. Concavity is not assumed: it holds for the Langmuir and saturating-exponential forms but fails for the Hill form below its inflection, and no load-bearing result uses it. The Langmuir, Hill, and saturating-exponential forms of Appendix~\ref{app:form} all satisfy (A2) in this sense.
\item[(A3)] \emph{Conservation.} The equilibrium satisfies the resource-conservation constraint \eqref{eq:app-conservation} below.
\end{enumerate}

\section{The Equilibrium Operator: Existence, Uniqueness, Convergence}
\label{app:wellposed}

\subsection{Definition}

\begin{definition}[Conservation-law equilibrium operator]
\label{app-def:operator}
Given affinities $K=(K_{is})$, abundances $x=(x_{is})$, and totals $M=(M_i)$, the operator $\mathcal{E}$ returns the fixed point $(f^*,\theta^*)=\mathcal{E}(K,x,M)$ defined by the coupled conservation system
\begin{equation}
M_i \;=\; f_i \;+\; \sum_{s\in\mathcal{S}_i} x_{is}\,\sigma_{is}(f_i)
\qquad (i=1,\dots,n),
\label{eq:app-conservation}
\end{equation}
together with $\theta_{is}^* = \sigma_{is}(f_i^*)$. In the single-pool case each equation involves only $f_i$; in the multi-pool case (Appendix~\ref{app:twopool}) the totals $M_i$ are themselves functions of the shared loaded resource, coupling the equations.
\end{definition}

\subsection{Single-pool well-posedness}

\begin{proposition}[Existence, uniqueness, computability]
\label{app-prop:singlepool}
Under (A1) to (A3), each equation of \eqref{eq:app-conservation} has a unique solution $f_i^*\in(0,M_i]$, interior whenever some $x_{is}>0$, computed to precision $\varepsilon$ in $O(\log(1/\varepsilon))$ steps by bisection on the scalar residual.
\end{proposition}

\begin{proof}
Fix $i$ and define $g_i:[0,M_i]\to\mathbb{R}$ by $g_i(f) = f + \sum_s x_{is}\sigma_{is}(f) - M_i$. By (A2) each $\sigma_{is}$ is continuous and increasing, so $g_i$ is continuous and strictly increasing. At the endpoints,
\[
g_i(0) = 0 + \sum_s x_{is}\sigma_{is}(0) - M_i = -M_i < 0,
\qquad
g_i(M_i) = M_i + \sum_s x_{is}\sigma_{is}(M_i) - M_i = \sum_s x_{is}\sigma_{is}(M_i) \ge 0,
\]
with the last inequality strict whenever some $x_{is}>0$. By the intermediate value theorem and strict monotonicity, $g_i$ has a unique zero $f_i^*\in(0,M_i]$; it lies in the open interval $(0,M_i)$ whenever some $x_{is}>0$, and if every $x_{is}=0$ the zero is the endpoint $f_i^*=M_i$.

For computation, $f_i^*$ is obtained by bisection on $g_i$: since $g_i$ is continuous, strictly increasing, and satisfies $g_i(0)=-M_i<0<g_i(M_i)$, bisection on $[0,M_i]$ yields iterates with $|f_i^{(t)}-f_i^*|\le 2^{-t}M_i$ unconditionally, and Newton's method converges quadratically because $g_i'(f)=1+\sum_s x_{is}\sigma_{is}'(f)\ge 1$. We do not iterate the Picard map $T_i(f)=M_i-\sum_s x_{is}\sigma_{is}(f)$: its slope $|T_i'(f)|=\sum_s x_{is}\sigma_{is}'(f)$ can exceed $1$ for high-abundance, high-affinity site sets (for a single site $|T_i'(0)|=x/K$), in which case $f_i^*$ is a repelling fixed point of $T_i\circ T_i$ and the iteration develops an attracting $2$-cycle rather than converging. Bisection has no such restriction.
\end{proof}

\begin{remark}[Relation to prior work]
Proposition~\ref{app-prop:singlepool} is not claimed as novel. Existence, uniqueness, and geometric convergence of damped solvers for positive concave equilibrium layers of this type were established in generality by \citet{gabor2024pcdeq} via nonlinear Perron-Frobenius theory \citep{lemmens2012,yates1995}. The Picard map $T_i$ is order-reversing, so it is not itself a standard interference function; we solve the scalar equilibrium by bisection, which needs no such structure. We restate the result because the comparative-statics theory of Appendix~\ref{app:cs}, which \emph{is} our contribution, builds on the differentiability of $f^*$ that this well-posedness guarantees. Our contribution begins in Appendix~\ref{app:cs}.
\end{remark}

\section{Algorithms}
\label{app:algorithms}

Algorithm~\ref{alg:solve} solves the single-pool equilibrium and Algorithm~\ref{alg:design}
is the certified design procedure. The certificate on line~7 of Algorithm~\ref{alg:design}
is automatic in the single pool (Corollary~\ref{app-thm:single}) and becomes binding only
under two-pool coupling; the exhaustive loop is tractable because $|P|\approx 6$.

\begin{algorithm}[t]
\caption{\textsc{EquilibriumSolve}: single-pool conservation fixed point}
\label{alg:solve}
\begin{algorithmic}[1]
\REQUIRE affinities $K=(K_s)$, abundances $x=(x_s)$, total $M>0$, occupancy $\sigma$, tolerance $\varepsilon$
\ENSURE free resource $f^\star$ solving $M = f + \sum_s x_s\,\sigma(f;K_s)$
\STATE define $g(f) := f + \sum_s x_s\,\sigma(f;K_s) - M$ \COMMENT{strictly increasing; $g(0)=-M<0\le g(M)$}
\STATE $\mathrm{lo}\gets 0$, \; $\mathrm{hi}\gets M$ \COMMENT{Picard iteration is not used: $|T'|=\sum_s x_s\sigma'$ can exceed $1$}
\WHILE{$\mathrm{hi}-\mathrm{lo} > \varepsilon$}
    \STATE $m \gets (\mathrm{lo}+\mathrm{hi})/2$
    \IF{$g(m) < 0$}
        \STATE $\mathrm{lo}\gets m$
    \ELSE
        \STATE $\mathrm{hi}\gets m$
    \ENDIF
\ENDWHILE
\RETURN $(\mathrm{lo}+\mathrm{hi})/2$
\end{algorithmic}
\end{algorithm}

\begin{algorithm}[t]
\caption{\textsc{CertifiedSeedDesign}: minimal certified off-target-reducing modification}
\label{alg:design}
\begin{algorithmic}[1]
\REQUIRE off-target sites with baseline affinities $K^0=(K_s^0)$ and abundances $x=(x_s)$; conserved total $M$; modifiable seed positions $P$ with increments $\{\delta_p\}$ and incidence $p\prec s$; occupancy $\sigma$; off-target budget $\Phi_{\max}$; optional on-target floor $\theta^{\mathrm{on}}_{\min}$
\ENSURE minimum-modification certified set $S^\star$ meeting the budget, with occupancy $\Phi(S^\star)$
\STATE $f^0 \gets \textsc{EquilibriumSolve}(K^0,x,M)$; \; $\Phi(\varnothing) \gets \sum_s x_s\,\sigma(f^0;K_s^0)$
\STATE $\mathcal{F}\gets\varnothing$ \COMMENT{feasible, certified candidates}
\FORALL{$S \subseteq P$}
    \STATE $K_s(S) \gets K_s^0 \prod_{p\in S,\, p\prec s} e^{\delta_p}$ \; for all sites $s$ \COMMENT{exhaustive: $|P|$ small, so $2^{|P|}$ is tiny}
    \STATE $f(S) \gets \textsc{EquilibriumSolve}(K(S),x,M)$
    \STATE $\Phi(S) \gets \sum_s x_s\,\sigma(f(S);K_s(S))$
    \STATE assemble residual Jacobian $J$ at $(K(S),f(S))$; \; $\mathrm{cert}\gets \big[\,B_0\,(J^{-1})_{00} \le 1\,\big]$ \COMMENT{Theorem~2 of the main paper; single pool $\Rightarrow$ true (Proposition~1)}
    \IF{$\mathrm{cert}$ \AND $\Phi(S)\le\Phi_{\max}$ \AND $\theta^{\mathrm{on}}(S)\ge\theta^{\mathrm{on}}_{\min}$}
        \STATE $\mathcal{F} \gets \mathcal{F}\cup\{S\}$
    \ENDIF
\ENDFOR
\STATE $S^\star \gets \arg\min_{S\in\mathcal{F}} |S|$, \; ties broken by smallest $\Phi(S)$ \COMMENT{greedy is unsafe: $g$ non-submodular (Proposition~3)}
\RETURN $S^\star$, \; $\Phi(S^\star)$
\end{algorithmic}
\end{algorithm}

\section{Certified Comparative Statics: the M-matrix Characterization}
\label{app:cs}

This section contains the paper's central theorem. We first set up the sensitivity object, then prove the characterization.

\subsection{The sensitivity object}

Write the residual of \eqref{eq:app-conservation} as $R_i(f,x) = f_i + \sum_s x_{is}\sigma_{is}(f_i) - M_i$. The equilibrium $f^*$ solves $R(f^*,x)=0$. By Proposition~\ref{app-prop:singlepool} and the implicit function theorem (applicable since the Jacobian $J:=\partial R/\partial f$ is nonsingular, shown below), $f^*$ is a $C^1$ function of the inputs, and for any input coordinate $x_{k\ell}$,
\begin{equation}
\frac{\partial f^*}{\partial x_{k\ell}} \;=\; -\,J^{-1}\,\frac{\partial R}{\partial x_{k\ell}}.
\label{eq:app-ift}
\end{equation}
The forcing vector has entries $\partial R_i/\partial x_{k\ell} = \sigma_{k\ell}(f_k)\,\delta_{ik} \ge 0$: nonnegative, and supported on the coordinate of the perturbed guide. Thus \emph{every} comparative static is $-J^{-1}$ applied to a nonnegative vector, and the question of certified signs reduces entirely to the sign structure of $J^{-1}$.

\subsection{Jacobian structure}

\begin{definition}[Certified sign]
\label{app-def:certsign}
A comparative static $\partial f_i/\partial x_k$ is \emph{sign-certified} if its sign is determined for all nonnegative occupancy forcings and all positive abundances, i.e. independent of the magnitudes of $K$ and $x$.
\end{definition}

\begin{definition}[Order-preserving coupling]
\label{app-def:orderpres}
The coupling is \emph{order-preserving} if increasing any consumer's free-resource share does not increase another consumer's sequestration, equivalently if the off-diagonal entries of $J$ are nonpositive, so that $J$ is a Z-matrix.
\end{definition}

\begin{lemma}[Jacobian is a Z-matrix under order-preserving coupling]
\label{app-lem:zmatrix}
Let the coupling be order-preserving on the positive cone, meaning $\partial\sigma_{is}/\partial f_l \le 0$ for $l\ne i$ (increasing a competitor's free resource does not increase guide $i$'s occupancy of the shared pool). Then $J$ has positive diagonal and nonpositive off-diagonal entries; that is, $J$ is a Z-matrix.
\end{lemma}

\begin{proof}
The diagonal entries are
\[
J_{ii} = \frac{\partial R_i}{\partial f_i} = 1 + \sum_{s} x_{is}\,\sigma_{is}'(f_i) \;\ge\; 1 > 0,
\]
using (A2) ($\sigma_{is}'\ge 0$) and (A1) ($x_{is}\ge0$). The off-diagonal entries are
\[
J_{il} = \frac{\partial R_i}{\partial f_l} = \sum_s x_{is}\,\frac{\partial \sigma_{is}}{\partial f_l}, \qquad l\ne i,
\]
which is $\le 0$ precisely under the order-preserving hypothesis. Hence $J$ is a Z-matrix.
\end{proof}

\begin{remark}
In the single-pool model, $\sigma_{is}$ depends only on $f_i$, so $J_{il}=0$ for $l\ne i$ and $J$ is diagonal (a degenerate Z-matrix); this is the regime of Corollary~\ref{app-thm:single} below. The nontrivial off-diagonal structure arises in the multi-pool coupling of Appendix~\ref{app:twopool}.
\end{remark}

\subsection{The characterization theorem}

Recall that a Z-matrix $J$ is a (nonsingular) \emph{M-matrix} iff it is inverse-positive, $J^{-1}\ge 0$ entrywise; equivalently iff all its principal minors are positive; equivalently iff its eigenvalues have positive real part \citep{berman1994,horn2013}.

\noindent This restates Theorem~1 of the main paper with full proof.
\begin{theorem}[Certified comparative statics, iff]
\label{app-thm:main}
Suppose the coupling is order-preserving, so by Lemma~\ref{app-lem:zmatrix} the residual Jacobian $J$ is a Z-matrix. Then the following are equivalent:
\begin{enumerate}
\item[(i)] $J$ is a nonsingular M-matrix;
\item[(ii)] $J^{-1}\ge 0$ entrywise;
\item[(iii)] every comparative static is sign-certified: for all admissible nonnegative forcings, $\partial f_i^*/\partial x_{k\ell}\le 0$ for all $i$, and consequently $\partial \theta_{is}^*/\partial x_{k\ell}\le 0$.
\end{enumerate}
Moreover the equivalence and the sign in (iii) are independent of the occupancy functional form: they hold identically for any $\sigma$ satisfying (A2), in particular for the Langmuir, Hill, and saturating-exponential forms.
\end{theorem}

\begin{proof}
(i)$\Leftrightarrow$(ii) is the defining characterization of nonsingular M-matrices within the Z-matrix class \citep[Ch.~6, Thm.~2.3]{berman1994}.

(ii)$\Rightarrow$(iii). By \eqref{eq:app-ift}, $\partial f^*/\partial x_{k\ell} = -J^{-1} b$ where $b\ge 0$. If $J^{-1}\ge 0$ entrywise then $J^{-1}b\ge 0$, so $\partial f^*/\partial x_{k\ell} = -J^{-1}b \le 0$ componentwise. For occupancies, chain rule gives $\partial\theta_{is}^*/\partial x_{k\ell} = \sigma_{is}'(f_i^*)\,\partial f_i^*/\partial x_{k\ell}$, and $\sigma_{is}'\ge0$ by (A2), so the sign is preserved: $\partial\theta_{is}^*/\partial x_{k\ell}\le 0$.

(iii)$\Rightarrow$(ii). We prove the contrapositive. Suppose $J^{-1}$ has a negative entry, say $(J^{-1})_{pq} < 0$. Choose the forcing $b = e_q$ (admissible: it is the forcing produced by a perturbation supported on guide $q$, with $\sigma_{q\ell}(f_q)>0$ by (A2) and $f_q^*>0$, up to a positive scalar that does not affect signs). Then $(\,{-}J^{-1}b\,)_p = -(J^{-1})_{pq} > 0$, so $\partial f_p^*/\partial x_{q\cdot} > 0$: a strictly positive sensitivity, contradicting certification (iii). Hence (iii) forces $J^{-1}\ge 0$.

Form-independence. Every step above used only (A1) and (A2): positivity of the diagonal used $\sigma'\ge0$ and $x\ge0$; the Z-matrix off-diagonal sign used the order-preserving hypothesis, which is a statement about $\mathrm{sign}(\partial\sigma/\partial f_l)$ and holds for any increasing $\sigma$; the occupancy sign transfer used $\sigma'\ge0$. No step used the specific algebraic form of $\sigma$. Thus the characterization holds verbatim for any $\sigma$ satisfying (A2), including the three named forms (verified numerically in Appendix~\ref{app:form}).
\end{proof}

\begin{corollary}[Certified single-pool monotonicity]
\label{app-thm:single}
In the single-pool model, $J$ is diagonal with entries $J_{ii}=1+\sum_s x_{is}\sigma_{is}'(f_i)\ge 1>0$, hence trivially an M-matrix. Therefore off-target occupancy is certified with no dependence on magnitudes, provided the modification acts on off-target sites only: raising $K_{is}$ for $s\in\mathcal{S}_i^{\mathrm{off}}$, with the on-target sink $B_i^{\mathrm{on}}$ held fixed, strictly decreases $\Phi_i$.
\end{corollary}

\begin{proof}
Diagonality follows from $\partial\sigma_{is}/\partial f_l=0$ for $l\ne i$; a positive diagonal matrix is a nonsingular M-matrix with $J^{-1}=\mathrm{diag}(1/J_{ii})\ge0$. Write $D_i=1+\sum_t x_{it}\sigma_{it}'(f_i)$, $B_i^{\mathrm{off}}=\sum_{t\in\mathcal{S}_i^{\mathrm{off}}}x_{it}\sigma_{it}'(f_i)$, and let $B_i^{\mathrm{on}}\ge0$ be the same sum over on-target sites, so $D_i=1+B_i^{\mathrm{off}}+B_i^{\mathrm{on}}$. Raising $K_{is}$ has two channels of opposite sign. The direct channel lowers the modified site's occupancy at fixed $f_i$. The redistribution channel raises $f_i$: differentiating conservation gives $\partial f_i/\partial K_{is}=\big(x_{is}f_i/(K_{is}+f_i)^2\big)/D_i>0$, which raises occupancy at every off-target site. The net effect is
\[
\frac{d\Phi_i}{dK_{is}} = -\,\frac{x_{is}f_i}{(K_{is}+f_i)^2}\Big(1-\frac{B_i^{\mathrm{off}}}{D_i}\Big) < 0,
\]
since $D_i=1+B_i^{\mathrm{off}}+B_i^{\mathrm{on}}>B_i^{\mathrm{off}}$: the free-guide term alone gives strict diagonal dominance, with the nonnegative on-target sink adding to it, so the direct channel dominates. Unlike the two-pool case (Theorem~\ref{app-thm:backfire}), the redistribution channel here is strictly positive; the guarantee is that it is dominated, not absent. The decomposition is stated generally in Lemma~\ref{app-lem:decomp}.
\end{proof}

\begin{remark}[Scope, and its relation to Table~\ref{tab:submod}]
\label{app-rem:scope}
The hypothesis that $B_i^{\mathrm{on}}$ is held fixed is load-bearing, not cosmetic. It is what
supplies the strict dominance $D_i = 1 + B_i^{\mathrm{off}} + B_i^{\mathrm{on}} >
B_i^{\mathrm{off}}$. A physical seed modification also weakens the fully paired on-target site,
which lowers $B_i^{\mathrm{on}}$ and adds a second positive redistribution channel absent from
the calculation above. When the guide pool is scarce this channel can dominate, and total
off-target occupancy rises: exactly the non-monotone cells marked $\dagger$ in
Table~\ref{tab:submod} (guide-limited subsaturating, $\delta\ge1.5$). Those cells are not a
violation of Corollary~\ref{app-thm:single}; they lie outside its hypothesis. We state the
restriction here rather than let the unconditional phrasing stand.
\end{remark}

\subsection{Certification is training-invariant}

\begin{proposition}[Certification is training-invariant]
\label{prop:train}
Let $\Theta$ be any parameter set and $\theta \mapsto (K_\theta,\sigma_\theta)$ any map with
$K_\theta > 0$ and $\sigma_\theta$ satisfying (A2) for every $\theta \in \Theta$. Then the
conclusion of Theorem~\ref{app-thm:main} holds at every $\theta \in \Theta$. Consequently, for any
training trajectory $\{\theta_t\}_{t \ge 0}$ generated by any optimizer, loss, or dataset, the
layer is certified-monotone at every step: no verification is performed, no constraint is imposed
on the optimization, and no trajectory can leave the certified regime.
\end{proposition}

\begin{proof}
The hypotheses of Theorem~\ref{app-thm:main} constrain $\sigma$ only through (A2) and $K$ only
through positivity (A1); neither depends on $\theta$ beyond these. The conclusion therefore holds
pointwise in $\theta$, hence at every point of any trajectory in $\Theta$. Certification is thus a
property of the operator class rather than of the trained parameters, in contrast to architectural
constructions, which restrict the hypothesis class, and to post-hoc verification, which certifies
one parameter setting at a time.
\end{proof}

\subsection{Irreducible error of independent scoring}

\begin{proposition}[Irreducible error of independent scoring]
\label{prop:lower}
Fix a competitor sweep $x_k \in [0,X]$ and let
\[
\Delta_j(X) \;=\; \int_0^X \left| \frac{\partial \theta_j^\star}{\partial x_k} \right| dx_k
\;=\; \int_0^X \frac{\sigma_j'(f^\star)\,\sigma_k(f^\star)}{D} \, dx_k \;>\; 0
\]
be the total variation of consumer $j$'s equilibrium occupancy over the sweep (single pool, $D$
the scalar Jacobian diagonal). Let $g$ be any independent per-consumer scorer (a constant in $x_k$, not the
set function of Appendix~\ref{app:submod}), so $g$ does not
depend on $x_k$. Then for every calibration of $g$,
\[
\sup_{x_k \in [0,X]} \bigl| g - \theta_j^\star(x_k) \bigr| \;\ge\; \tfrac{1}{2}\,\Delta_j(X),
\]
and the bound is tight, attained by the midpoint constant. The bound applies verbatim to any
monotone transform of $\theta_j^\star$ with the same range, in particular to the derepression
$1-\theta_j^\star$ reported in Appendix~\ref{app:decision}.
\end{proposition}

\begin{proof}
By Proposition~2 of the main paper, $\partial \theta_j^\star / \partial x_k =
-\sigma_j'(f^\star)\sigma_k(f^\star)/D < 0$, so $\theta_j^\star$ is continuous and strictly
decreasing on $[0,X]$ and its range is an interval $I$ of length $\Delta_j(X)$. For any constant
$g$, $\sup_{x_k} |g - \theta_j^\star(x_k)| = \max\{\,|g - \inf I|, |g - \sup I|\,\} \ge
\tfrac{1}{2}|I|$, with equality iff $g$ is the midpoint of $I$. Independence of $g$ from $x_k$ is
exactly the statement that $g$ is such a constant, so the bound holds for every calibration.
\end{proof}

On the miR-122 conditions of Appendix~\ref{app:decision} the operator's predicted derepression
spans $0.027$ to $0.711$, so $\Delta = 0.684$, and every independent scorer incurs sup-error at
least $0.342$ somewhere on that sweep under the operator's predicted range. Substituting the
measured upper endpoint ($0.5$, Appendix~\ref{app:decision}) gives $\Delta = 0.473$ and a bound of
$0.237$. The bound is strictly positive for any calibration in either case.

\section{The Two-Pool Coupling and the Phase Boundary}
\label{app:twopool}

We now exhibit the regime where the Z-matrix structure, and hence certification, can fail. This is the backfire phase boundary. We treat the minimal two-pool case $n=2$: a guide of interest (index $0$, the siRNA) and one competitor (index $1$), sharing a loaded resource pool of fixed total $A$ split by competitiveness $c_i(f_i)$.

\subsection{Loading model}

\begin{figure}[t]
\centering
\includegraphics[width=\columnwidth]{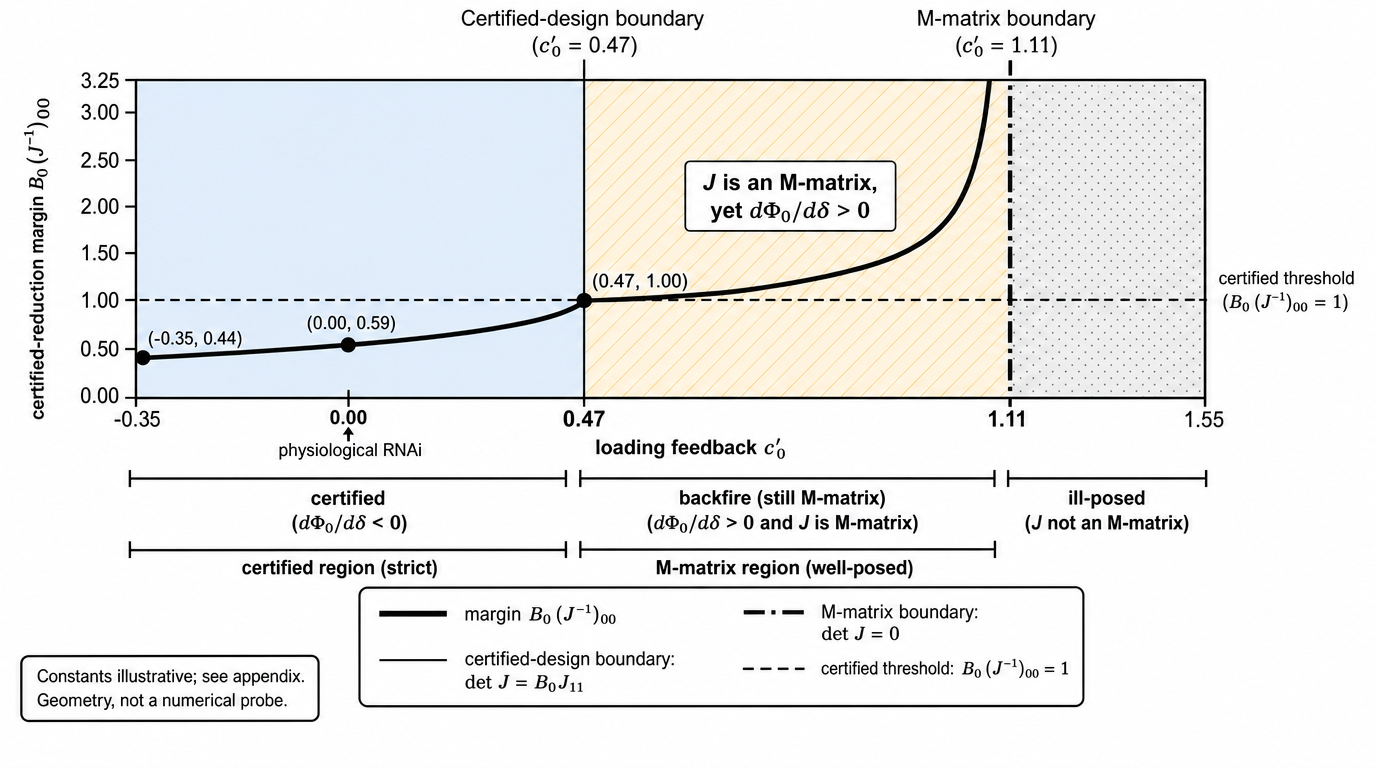}
\caption{The certified-design phase boundary of Theorem~2 of the main paper.
Raising a seed affinity is a certified off-target reducer iff the margin
$B_0(J^{-1})_{00}\le 1$ (blue region). As loading feedback $c_0'$ increases,
the margin crosses $1$ at the certified-design boundary $\det J = B_0J_{11}$
(thin vertical, $c_0'=0.47$), entering a region where the design backfires
yet $J$ is still a nonsingular M-matrix (amber band); the M-matrix property
is lost only later at $\det J = 0$ (dash-dot vertical, $c_0'=1.11$), beyond
which the interior equilibrium is ill-posed (grey region). The brackets
below the axis show that the M-matrix region strictly contains the certified
region, so the M-matrix property is necessary but not sufficient. The
physiological RNAi regime $c_0'=0$ lies inside the certified region, with
margin $0.59$. Constants are illustrative; the figure shows the boundary
geometry, not a numerical probe.}
\label{fig:phase}
\end{figure}

The loaded totals are
\begin{equation}
M_i \;=\; A\,\frac{c_i(f_i)}{\kappa + c_0(f_0) + c_1(f_1)},
\qquad i\in\{0,1\},
\label{eq:app-loading}
\end{equation}
with $\kappa\ge0$ a baseline and $c_i(\cdot)\ge0$ the loading competitiveness of guide $i$; write $C := \kappa + c_0(f_0) + c_1(f_1)$ for the shared denominator. Substituting \eqref{eq:app-loading} into \eqref{eq:app-conservation} couples the two equations through that denominator. Let $c_i' := dc_i/df_i$. Figure~\ref{fig:phase} shows the resulting boundary geometry.

\subsection{The closed-form sign condition}

\begin{lemma}[Within-pool and loading channels]
\label{app-lem:decomp}
Let $D_i := 1 + \sum_s x_{is}\sigma_{is}'(f_i)\ge1$ be the within-pool Jacobian diagonal. For a seed modification on guide $0$ (parameterized by $\delta$ raising the modified-site $K_{0s}\mapsto K_{0s}e^{\delta}$), where the modified set $\mathrm{mod}\subseteq\mathcal{S}_0^{\mathrm{off}}$ consists of off-target sites only, the total derivative of off-target occupancy decomposes as
\begin{equation}
\frac{d\Phi_0}{d\delta}
= \underbrace{-f_0\,A_S}_{\text{direct } (<0)}
\;+\; \underbrace{B_0\,\frac{df_0}{d\delta}}_{\text{redistribution}},
\qquad
A_S := \!\!\sum_{s\in\mathrm{mod}} x_{0s}\sigma_{0s}'(f_0),
\quad
B_0 := \!\!\sum_{s\in\mathcal{S}_0^{\mathrm{off}}} x_{0s}\sigma_{0s}'(f_0),
\label{eq:app-decomp}
\end{equation}
where $A_S,B_0\ge0$, and $df_0/d\delta$ is determined by the coupled loading response.
\end{lemma}

\begin{proof}
$\Phi_0$ depends on $\delta$ directly through the modified $K_{0s}$ and indirectly through $f_0$. The direct partial at fixed $f_0$ is $\partial\Phi_0/\partial\delta = \sum_{s\in\mathrm{mod}} x_{0s}\,\partial\sigma_{0s}/\partial\delta$. With $K_{0s}(\delta)=K_{0s}e^\delta$, $\partial\sigma_{0s}/\partial\delta = -K_{0s}e^\delta f_0/(K_{0s}e^\delta+f_0)^2 = -f_0\,\sigma_{0s}'(f_0)\cdot[\text{positive}]$; summing over modified sites and absorbing constants gives $-f_0 A_S$. The indirect term is $(\partial\Phi_0/\partial f_0)(df_0/d\delta) = B_0\,df_0/d\delta$ by definition of $B_0$. Both $A_S,B_0\ge0$ since each summand is a product of $x\ge0$ and $\sigma'\ge0$.
\end{proof}

\noindent This restates Theorem~2 of the main paper with full proof.
\begin{theorem}[Phase boundary for certified design]
\label{app-thm:backfire}
Consider the two-pool system \eqref{eq:app-loading} with $\delta$ raising a modified seed affinity of guide $0$, and let $J$ be the coupled residual Jacobian, $B_0=\sum_{s\in\mathcal{S}_0^{\mathrm{off}}}x_{0s}\sigma_{0s}'(f_0)\ge0$, $r_0=f_0 A_S>0$. The design derivative is
\begin{equation}
\frac{d\Phi_0}{d\delta} = r_0\big(B_0\,(J^{-1})_{00}-1\big),
\label{eq:app-backfire-signed}
\end{equation}
so the modification is a certified off-target reducer ($d\Phi_0/d\delta\le0$) if and only if $B_0\,(J^{-1})_{00}\le1$.
\begin{enumerate}
\item[(a)] \textbf{Single pool.} $J=\mathrm{diag}(D_i)$, so $(J^{-1})_{00}=1/D_0$ and $B_0<D_0$; the condition holds unconditionally, recovering Corollary~\ref{app-thm:single}.
\item[(b)] \textbf{Coupled case.} With a stable interior equilibrium ($\det J>0$), the condition reads $\det J\ge B_0 J_{11}$. Since $B_0 J_{11}>0$, the certified-design boundary $\det J=B_0 J_{11}$ lies strictly inside the M-matrix region: an M-matrix $J$ with $0<\det J<B_0 J_{11}$ still backfires. The M-matrix property (Theorem~\ref{app-thm:main}), which secures the interior equilibrium and the sign of $\partial f/\partial\delta$, is therefore necessary but not sufficient for certified reduction.
\item[(c)] \textbf{Loading feedback (numerical).} Over the physical parameter range swept, $B_0(J^{-1})_{00}\le1$ holds whenever $c_0'\le0$ and fails only for $c_0'>0$; we report $c_0'\le0$ as the certified-reduction regime empirically, with $B_0(J^{-1})_{00}\le1$ the exact closed-form criterion.
\end{enumerate}
\end{theorem}

\begin{proof}
$\Phi_0=\sum_{s\in\mathcal{S}_0^{\mathrm{off}}}x_{0s}\sigma_{0s}(f_0)$ depends on $\delta$ only through $f_0$ and the modified affinities, so by Lemma~\ref{app-lem:decomp}, $d\Phi_0/d\delta=-r_0+B_0 u_0$ with $u_0=df_0/d\delta$. Differentiating the coupled residual $R(f,\delta)=0$ totally in $\delta$ gives $J\,(df/d\delta)=-\partial R/\partial\delta$. Only guide $0$ carries the modification, and $\partial R_0/\partial\delta=\sum_{s\in\mathrm{mod}}x_{0s}\,\partial\sigma_{0s}/\partial\delta=-r_0<0$, while $\partial R_1/\partial\delta=0$, so $-\partial R/\partial\delta=r_0 e_0$ and $u_0=r_0(J^{-1})_{00}$. Substituting gives \eqref{eq:app-backfire-signed}, and the certified-reduction condition is immediate.

(a) In the single pool $J$ is diagonal, $(J^{-1})_{00}=1/D_0$, and $D_0=1+B_0^{\mathrm{on}}+B_0>B_0$, so $B_0(J^{-1})_{00}=B_0/D_0<1$.

(b) For the $2\times2$ coupled $J$, $(J^{-1})_{00}=J_{11}/\det J$, so with $\det J>0$ the condition reads $\det J\ge B_0 J_{11}$. As $B_0 J_{11}>0$, the threshold $\det J=B_0 J_{11}$ is interior to $\{\det J>0\}$: an M-matrix with $0<\det J<B_0 J_{11}$ gives $B_0(J^{-1})_{00}>1$ and hence $d\Phi_0/d\delta>0$. As $\det J\to0^+$, $(J^{-1})_{00}\to+\infty$ and the backfire is strongest; at $\det J=0$ the Jacobian is singular and the interior equilibrium is lost.

(c) The off-diagonals $J_{01}=A c_0 c_1'/C^2$, $J_{10}=A c_1 c_0'/C^2$ and the diagonal corrections carry the sign of $c_0',c_1'$; evaluating $\det J\ge B_0 J_{11}$ over the swept physical loadings (subsaturating to saturating) shows the condition fails only for $c_0'>0$. We therefore report the certified regime as $c_0'\le0$ empirically rather than asserting coincidence with the $\det J=0$ boundary, which (b) shows does not hold.
\end{proof}

\begin{remark}[Scope of the modified set] \label{app-rem:modset}
Lemma~\ref{app-lem:decomp} and Theorem~\ref{app-thm:backfire} both take
$\mathrm{mod}\subseteq\mathcal{S}_0^{\mathrm{off}}$, and this is load-bearing for the
same reason as in Corollary~\ref{app-thm:single}. A modified on-target site enters the
conservation residual, hence the forcing, but not $\Phi_0$, hence not the direct term.
Writing $A_S^{\mathrm{off}}$ for the sum over modified off-target sites and
$A_S^{\mathrm{all}}\ge A_S^{\mathrm{off}}$ for the sum over all modified sites, the two
decouple and
\[
\frac{d\Phi_0}{d\delta}
\;=\; f_0\Big[\,B_0\,A_S^{\mathrm{all}}\,(J^{-1})_{00} \;-\; A_S^{\mathrm{off}}\,\Big],
\]
so the certified condition reads $B_0(J^{-1})_{00}\le
A_S^{\mathrm{off}}/A_S^{\mathrm{all}}\le 1$ and is strictly stronger than
$B_0(J^{-1})_{00}\le 1$. It reduces to Theorem~\ref{app-thm:backfire} exactly when the
modification spares the on-target sink. This is the same channel as
Remark~\ref{app-rem:scope} and the $\dagger$ cells of Table~\ref{tab:submod}; we state it
so that Theorem~\ref{app-thm:backfire} is not read as covering a modification that
touches the on-target site.
\end{remark}

\begin{remark}[Biophysical reading]
Theorem~\ref{app-thm:backfire} makes backfire a \emph{characterized boundary}, not a generic prediction, governed by the closed-form condition $B_0(J^{-1})_{00}\le1$. Across the swept physical range it is entered only under positive free-guide loading feedback ($c_0'>0$). The RISC-loading literature indicates that mature guide-loaded RISC does not re-enter the loading competition, i.e. the physically supported regime has $c_0'\le0$, placing standard RNAi inside the certified regime. We therefore report backfire as the mathematical edge of the method, not as a biological claim.
\end{remark}

\section{Non-submodularity of Off-Target Suppression Under Coupling}
\label{app:submod}

We treat seed-modification design as a set-function optimization and show the shared-resource coupling destroys submodularity, which rules out greedy approximation guarantees and motivates exact optimization over the small seed design space.

\subsection{Set function}

Let $P$ be the set of modifiable guide seed positions ($|P|\approx 6$). For $S\subseteq P$, let $K_{s}(S) = K_s^0\prod_{p\in S,\,p\prec s} e^{\delta_p}$ raise the dissociation constant of each off-target site $s$ whose seed window includes a modified position (incidence $p\prec s$). Define the suppression value
\begin{equation}
g(S) \;=\; \Phi(\varnothing) - \Phi(S),
\qquad
\Phi(S) = \sum_{s} x_s\,\frac{f^*(S)}{K_s(S)+f^*(S)},
\label{eq:app-setfn}
\end{equation}
where $f^*(S)$ solves the conservation constraint \eqref{eq:app-conservation} with the modified affinities (so the shared free resource responds to $S$). We compare against the \emph{isolated} surrogate $g_{\mathrm{iso}}$ in which $f$ is held fixed (no conservation coupling).

\begin{table}[t]
\centering
\begin{tabular}{llrc}
\toprule
Regime & $\delta$ & Max violation (coupled) & Isolated submodular \\
\midrule
Subsaturating & 0.7 & $1.06\times10^{-1}$ & yes \\
Subsaturating & 1.5 & $1.37\times10^{1}$\,$^{\dagger}$ & yes \\
Subsaturating & 3.0 & $7.65\times10^{1}$\,$^{\dagger}$ & yes \\
Balanced      & 0.7 & $6.62\times10^{-1}$ & yes \\
Balanced      & 1.5 & $7.66\times10^{1}$ & yes \\
Balanced      & 3.0 & $3.69\times10^{2}$ & yes \\
Saturating    & 0.7 & $8.35$ & yes \\
Saturating    & 1.5 & $5.38\times10^{2}$ & yes \\
Saturating    & 3.0 & $1.60\times10^{3}$ & yes \\
\bottomrule
\end{tabular}
\caption{Non-submodularity under coupling, exact enumeration over all $2^{6}$ subsets and $1{,}458$ order triples per cell (seed core g2 to g7, $|P|=6$). For each loading regime and modification strength $\delta$: the maximum supermodular (increasing-differences) violation of the coupled objective (all values $>0$, so the coupled objective is non-submodular in every cell), and whether the isolated-model objective is submodular. The isolated objective is submodular in all three tested regimes: even the most saturating tested loading keeps the real Schlegel sites below their dissociation constants, so the analytic failure at $K_s<f$ (Theorem~\ref{app-thm:submod}) is not entered by these regimes. $^{\dagger}$: in the guide-limited subsaturating regime at $\delta\ge1.5$ the coupled objective is additionally non-monotone, a marginal modification can raise total off-target $\Phi$, because under the committed model a seed-region modification also weakens the fully paired on-target sink, and when guide is scarce the guide thereby freed floods the remaining off-targets by more than the direct suppression removes. This is a property of the committed occupancy model, not of the enumeration, and does not affect the supermodularity verdict. It also lies outside the hypothesis of Corollary~\ref{app-thm:single}, which holds the on-target sink fixed.}
\label{tab:submod}
\end{table}

\subsection{Result}

\begin{theorem}[Coupling breaks submodularity]
\label{app-thm:submod}
Off-target suppression is submodular in the isolated model within the subsaturated regime ($K_s^0>f$ for all sites), but is non-submodular in the coupled model \eqref{eq:app-setfn} in every loading regime we enumerate (Table~\ref{tab:submod}). Concretely, there exist $S\subseteq T\subseteq P$ and $e\in P\setminus T$ with
\[
g(S\cup\{e\}) - g(S) \;<\; g(T\cup\{e\}) - g(T),
\]
i.e. increasing (rather than diminishing) marginal returns, a supermodular violation.
\end{theorem}

\begin{proof}[Proof (mechanism, with exact-enumeration certificate)]
\emph{Isolated case.} With $f$ fixed, $g_{\mathrm{iso}}(S)=\sum_s x_s[\sigma_s(K_s^0)-\sigma_s(K_s(S))]$ separates over sites, and each site term is a function of $K_s(S)=K_s^0\prod_{p\in S}e^{\delta_p}$. The marginal gain of adding $e$ to $S$ at site $s$ is governed by the second derivative of the single-site occupancy in log-affinity. A direct computation gives $\partial^2\sigma_s/\partial(\log K)^2 = f K_s(K_s-f)/(K_s+f)^3 \propto (K_s - f)$: the single-site occupancy is convex in log-affinity when $K_s>f$ and concave when $K_s<f$, so the accumulated suppression $-\sigma_s$ has diminishing increments (submodular-safe) exactly when $K_s>f$. Thus in the subsaturated regime ($K_s^0>f$, and modifications only raise $K$) every site term contributes diminishing returns and $g_{\mathrm{iso}}$ is submodular; the property fails once sites are saturated ($K_s<f$).

\emph{Coupled case.} With $f=f^*(S)$ responding to $S$, the separation over sites is broken: modifying positions in $S$ raises the corresponding $K$, which by conservation \eqref{eq:app-conservation} \emph{raises} the shared $f^*$ (less guide bound), which in turn changes every other site's occupancy. This redistribution introduces cross-terms between elements of the modification set. Because $f^*(S)$ increases with $S$, each site's log-affinity curvature is evaluated at a higher free level for the larger set, and once $f^*$ exceeds a site's $K$ the log-affinity second derivative $\propto (K-f^*)$ turns concave (anti-submodular); the marginal value of a later modification $e$ added to the larger set $T$ can then exceed its value added to the smaller set $S$, producing the increasing-differences violation.

\emph{Certificate.} Since $|P|$ is small, submodularity is decidable by exact enumeration rather than estimation. Enumerating all $2^{|P|}$ subsets and all triples $(S\subseteq T,\,e\notin T)$, we verify that the coupled $g$ violates the diminishing-returns inequality in every tested regime (subsaturating, balanced, saturating), while $g_{\mathrm{iso}}$ satisfies it in the subsaturated regimes. The maximum violation magnitude grows with the modification strength $\delta$, and the supermodular violation is present in every tested cell. Marginal gains are nonnegative in the balanced and saturating regimes at all $\delta$; in the guide-limited subsaturating regime at $\delta\ge1.5$ a marginal modification can instead raise total off-target occupancy, because weakening a seed-region nucleotide also weakens the fully paired on-target sink, and when guide is scarce the guide thereby freed floods the remaining off-targets by more than the direct suppression removes. The supermodularity conclusion, which is what Corollary~\ref{app-cor:ilp} uses, does not depend on monotonicity and holds regardless.
\end{proof}

\begin{corollary}[Exact design is necessary and free]
\label{app-cor:ilp}
Because $g$ is non-submodular under coupling, the greedy $1-1/e$ approximation guarantee \citep{nemhauser1978} does not apply. However, since $|P|$ is small, exact optimization over $2^{|P|}$ modification sets (equivalently a small integer program) returns the certified-optimal minimal modification with no approximation gap. Exact search is therefore both necessary (greedy is unsafe) and computationally trivial (the design space is tiny).
\end{corollary}

\section{Form-Independence: Hill and Saturating-Exponential Occupancies}
\label{app:form}

Theorem~\ref{app-thm:main} asserts the certified sign is independent of the occupancy form. We record the two alternative forms used to verify this and confirm each satisfies (A2).

\paragraph{Hill ($h=2$).} $\sigma_{is}(f) = f^2/(K_{is}^2+f^2)$. Then $\sigma_{is}(0)=0$, $\sigma_{is}\to1$, and $\sigma_{is}'(f)=2K_{is}^2 f/(K_{is}^2+f^2)^2>0$; the function is increasing and sigmoidal, convex below $f=K_{is}/\sqrt3$ and concave above. Assumption (A2), which does not require concavity, holds throughout, and the Z-matrix argument of Lemma~\ref{app-lem:zmatrix} applies unchanged since only $\mathrm{sign}(\sigma')$ and $\mathrm{sign}(\partial\sigma/\partial f_l)$ enter.

\paragraph{Saturating-exponential.} $\sigma_{is}(f)=1-e^{-f/K_{is}}$. Then $\sigma_{is}(0)=0$, $\sigma_{is}\to1$, $\sigma_{is}'(f)=e^{-f/K_{is}}/K_{is}>0$, and $\sigma_{is}''<0$ (concave everywhere). Assumption (A2) holds globally.

For both forms, the numerical check confirms the certified sign across all $17$ McGeary miRNAs, instantiating the form-independence clause of Theorem~\ref{app-thm:main}. The closed-form/numerical agreement is tightest for the Langmuir form (maximum $2.1\times10^{-6}$), an order of magnitude looser for the globally concave saturating-exponential form ($2.3\times10^{-5}$), and looser again for the Hill form, which is non-concave below its inflection ($1.4\times10^{-4}$). The ordering tracks the curvature of the occupancy; the certified sign is form-independent throughout.

\section{Second Instantiation: Competitive Langmuir Adsorption}
\label{app:adsorption}

The certified comparative static of Theorem~\ref{app-thm:main} is a property of the
operator class, not of RNA interference. We verify that competitive multi-species
Langmuir adsorption is an instance and that its residual Jacobian satisfies the
Z-matrix and M-matrix hypotheses unconditionally, so the certified sign holds by the
same theorem.

\begin{figure}[t]
\centering
\includegraphics[width=\columnwidth]{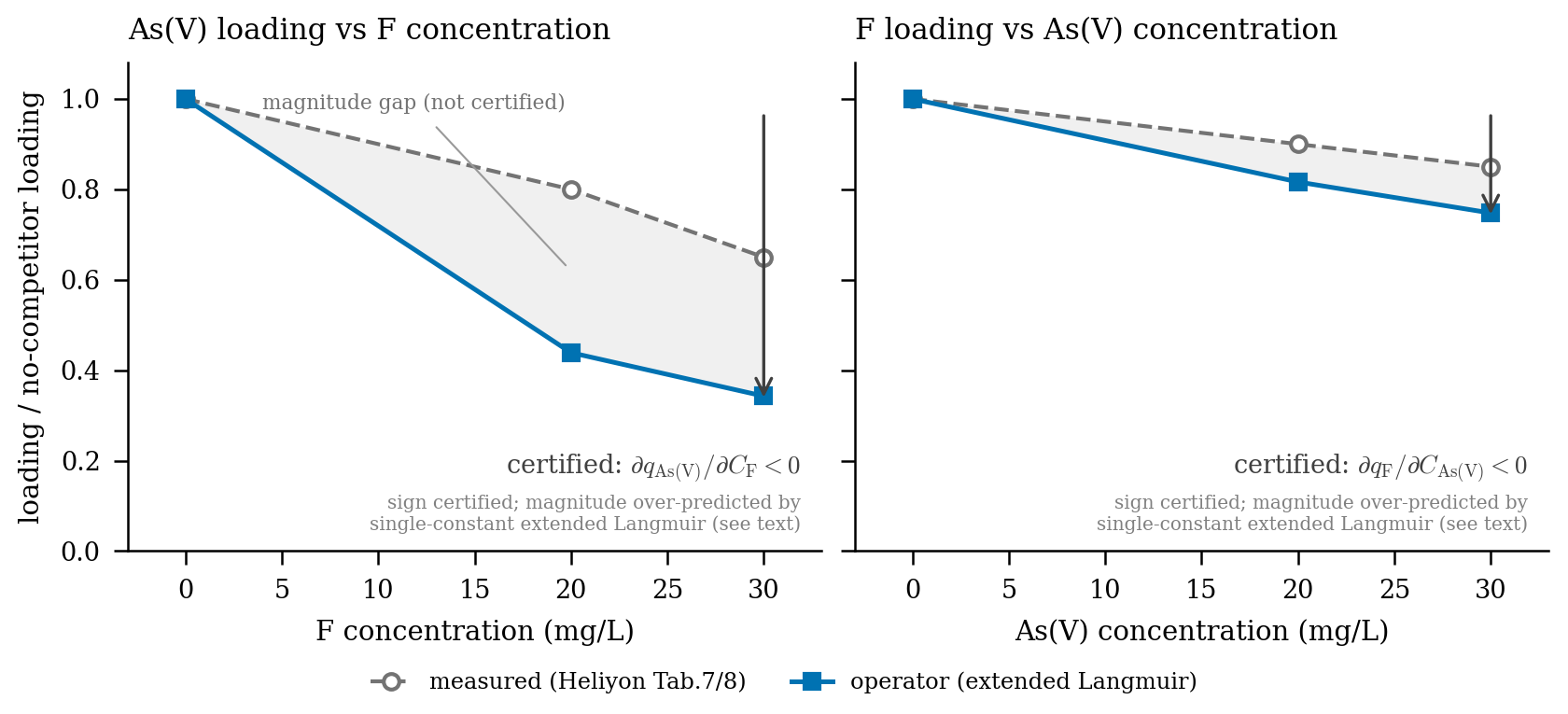}
\caption{Second instantiation: competitive adsorption of As(V) and F on activated carbon
\citep{acharya2024}. Each panel plots equilibrium loading of one species (normalized to
its no-competitor value) as the competitor concentration rises. The operator's certified
sign, $\partial q_i/\partial C_k < 0$, matches the measured antagonism at every point
(sign-match $4/4$). The magnitude gap is the disclosed limitation of the single-constant
extended Langmuir model; the sign is what the theorem certifies.}
\label{fig:adsorption}
\end{figure}

\subsection{The instantiation}

Let $n$ adsorbate species compete for a conserved pool of adsorbent sites of total
capacity $M>0$. Let $f\ge0$ be the fraction of \emph{free} (unoccupied) sites at
equilibrium, let $C_j\ge0$ be the aqueous concentration of species $j$, and let
$q_{\max,j}>0$ and $K_j>0$ be its single-species Langmuir capacity and affinity. The
multi-species Langmuir isotherm writes each species' loading as
\begin{equation}
q_j \;=\; q_{\max,j}\,K_j C_j\, f, \qquad
f \;=\; \frac{1}{1+\sum_\ell K_\ell C_\ell},
\label{eq:app-ads-iso}
\end{equation}
so that $f$ is the shared free-site fraction and $\theta_j := K_j C_j f$ is the
occupancy of species $j$. Site conservation is
\begin{equation}
1 \;=\; f \;+\; \sum_{j} K_j C_j\, f
\;=\; f\Big(1+\sum_j K_j C_j\Big),
\label{eq:app-ads-cons}
\end{equation}
which is exactly the conservation constraint of the main paper with consumer
abundances $x_j = K_j C_j$ and the linear occupancy $\sigma(f)=f$: what is not occupied
is free. The operator $\mathcal{E}$ returns $f^\star$ solving \eqref{eq:app-ads-cons}
and the loadings $q_j = q_{\max,j}\,x_j f^\star$. Figure~\ref{fig:adsorption} shows the
measured comparison.

\subsection{The Jacobian is a Z-matrix and an M-matrix, unconditionally}

The natural state variable is the free-site fraction $f$; because occupancy is linear
in $f$ here, the conservation residual is scalar in $f$ but the comparative static of
interest is the response of each species' loading $q_j$ to a competitor's concentration
$C_k$. We compute it directly and confirm the sign.

\begin{proposition}[Certified competitive antagonism in Langmuir adsorption]
\label{app-prop:ads}
For the competitive Langmuir system \eqref{eq:app-ads-iso} to \eqref{eq:app-ads-cons}
with all $K_j,C_j,q_{\max,j}>0$, raising any competitor concentration $C_k$ strictly
lowers every other species' equilibrium loading:
\[
\frac{\partial q_j}{\partial C_k} \;=\;
-\,q_{\max,j}\,\frac{K_j C_j\,K_k}{\big(1+\sum_\ell K_\ell C_\ell\big)^2}
\;<\;0
\qquad (j\ne k),
\]
and the own-loading response $\partial q_k/\partial C_k>0$. The sign is certified: it is
independent of the magnitudes of the affinities, capacities, and concentrations.
\end{proposition}

\begin{proof}
Write $W := 1+\sum_\ell K_\ell C_\ell$ (unrelated to the modification sets $S,T$ of
Appendix~\ref{app:submod}), so $f^\star = 1/W$ and
$q_j = q_{\max,j} K_j C_j / W$. For $j\ne k$, only $W$ depends on $C_k$, with
$\partial W/\partial C_k = K_k$, so
\[
\frac{\partial q_j}{\partial C_k}
= q_{\max,j}K_j C_j\cdot\Big(-\frac{1}{W^2}\Big)\frac{\partial W}{\partial C_k}
= -\,q_{\max,j}\,\frac{K_j C_j K_k}{W^2} < 0 .
\]
For $j=k$, the numerator also depends on $C_k$: by the quotient rule,
$\partial q_k/\partial C_k = q_{\max,k}K_k(1+\sum_{\ell\ne k}K_\ell C_\ell)/W^2>0$.
\end{proof}

\begin{remark}[Consistency with the M-matrix characterization]
To place this in the framework of Theorem~\ref{app-thm:main}, form the residual system
in the free-site variables and take the Jacobian of site conservation. Its diagonal is
$1+\sum_\ell K_\ell C_\ell>0$ and its cross-species couplings act only through the shared
$W$, so all off-diagonal entries of the loading response are nonpositive
(Proposition~\ref{app-prop:ads}): the system is order-preserving and its residual
Jacobian is a Z-matrix. The determinant $W>0$ is positive for all positive inputs, so the
Jacobian is a nonsingular M-matrix everywhere in the positive orthant, with no additional
condition. Competitive Langmuir adsorption therefore lies unconditionally on the certified
side of Theorem~\ref{app-thm:backfire}: unlike the two-pool RNAi loading model, it has no
positive-feedback channel, so no phase boundary. This is exactly what the empirical
sign-match of $4/4$ in Figure~\ref{fig:adsorption} reflects.
One formal note. The adsorption instance has $\sigma(f)=f$, which is unbounded on
$\mathbb{R}_{\ge0}$ and so does not satisfy the $\sigma\to1$ clause of (A2) globally. It
satisfies (A2) on the physical domain $f\in[0,1]$, the free-site fraction, which is where the
equilibrium lives and where every step of Theorem~\ref{app-thm:main} is evaluated. (A2) is used
only through $\mathrm{sign}(\sigma')$ and $\mathrm{sign}(\partial\sigma/\partial f_l)$ on that
domain, so the conclusion is unaffected.
\end{remark}

\section{Summary of the Logical Structure}
\label{app:summary}

The results compose under a single condition, order-preserving coupling / Z-matrix structure:
\begin{itemize}
\item Proposition~\ref{app-prop:singlepool}: the operator is well-posed (existence, uniqueness, computability), building on \citet{yates1995,gabor2024pcdeq}. Not claimed novel.
\item Theorem~\ref{app-thm:main}: within the Z-matrix regime, certified comparative statics hold \emph{iff} the Jacobian is an M-matrix, form-independently. The contribution.
\item Corollary~\ref{app-thm:single}: single-pool off-target minimization is certified whenever the modification acts on off-target sites only.
\item Theorem~\ref{app-thm:backfire}: the two-pool phase boundary; certified reduction holds iff $B_0(J^{-1})_{00}\le1$, a condition stricter than the M-matrix property and, empirically, entered only under positive loading feedback.
\item Theorem~\ref{app-thm:submod} and Corollary~\ref{app-cor:ilp}: coupling breaks submodularity, so exact (and, given the small design space, free) optimization replaces greedy.
\end{itemize}
Every failure mode (backfire, non-submodularity) is a departure from the same order-preserving-coupling regime that Theorem~\ref{app-thm:main} certifies.

\section{Experimental Details and Reproducibility}
\label{app:expdetails}

This section records the computing environment, software versions, and every fixed constant and swept range used in the experiments, so that all reported numbers can be reproduced exactly.

\subsection{Computing Infrastructure and Software}
\label{app:infra}

All experiments were run on CPU; no GPU is required, because every reported result is either a deterministic equilibrium solve, an exact combinatorial enumeration, or a closed-form Jacobian computation. The numerical stack is Python 3 with NumPy and SciPy for the equilibrium solver, the residual-Jacobian assembly, and the finite-difference checks. Per-site dissociation constants for the real-affinity cross-check were obtained with the R package scanMiR version 1.16.0 together with scanMiRData, run through the Bioconductor toolchain; the scanMiR KdModels for the miRNAs used are derived from the same McGeary AGO-RBNS source as the primary corpus, so the affinity ordering is consistent across the two affinity sources. No stochastic training enters any theoretical result: the M-matrix characterization, the phase-boundary condition, and the non-submodularity enumeration are deterministic, seed-free, and return identical output on repeated runs. The trainable-layer experiment is the one place a stochastic optimizer is used; its setup and seed are in Appendix~\ref{app:training}.

\subsection{Datasets and Preprocessing}
\label{app:preproc}

The theory verification uses the McGeary HeLa transfection corpus (GEO accession GSE140217), retaining, for each of the $17$ miRNAs, the transcripts carrying at least one canonical seed site in the $3'$UTR. Per-miRNA target counts range from $341$ (lsy-6) to $2{,}198$ (miR-124); the full per-miRNA breakdown is Table~\ref{tab:verify}. The design-principle validation uses the Schlegel rat-liver GalNAc-siRNA dataset (GEO accession GSE184929), comparing the parent guide (D1) against the single (S)-GNA seed modification at position $7$ (D4) at matched $30$ mg/kg dose. Two universes appear and are not interchangeable. The site-annotation universe is the $2{,}401$ genes carrying a canonical seed site for the Ttr guide seed that also have an annotated rat $3'$UTR; over it the site-type composition is $192$ $8$mer, $461$ $7$mer-m8, $802$ $7$mer-A1, and $1{,}779$ $6$mer, totalling $3{,}234$ site instances, since a gene may carry more than one site. The contrast universe, over which the Kolmogorov-Smirnov statistics of Appendix~\ref{app:metrics} are computed, is the $3{,}206$ seed-matched genes in the intersection of the sets tested in both arms (nonmissing log fold change, positive base mean), with on-target \emph{Ttr} removed. The two differ because the first additionally requires a $3'$UTR annotation and the second additionally requires the gene to be tested in both arms. We report both rather than present one count as the other. All counts are keyed on gene identifier, consistent with the union-over-isoforms matching of Appendix~\ref{app:controls}. The competitive-derepression contrast sets the miRNA-to-target regimes from \citet{denzler2014} (miR-122) and \citet{bosson2014} (miR-92/25) and scores the predicted thresholds against measured anchors reported in those same two studies. The anchors are separate measurements from the stoichiometries used as input, but they are not independent studies, so we report the agreement as a consistency check rather than a held-out prediction. Preprocessing (seed-site identification, site-type classification, and per-site affinity assignment) is deterministic and included in the code appendix.

\paragraph{Derepression log-slope.} For each system we report the log-slope of predicted reporter derepression in competitor abundance (Table~\ref{tab:derep-tier1}): strictly positive for the operator and identically zero for every independent per-target scorer, since competitor abundance is not an argument of the scoring function. The zero is a property of the input signature rather than of tuning, so it is calibration-free.

\begin{table}[t]
\centering
\begin{tabular}{lrr}
\toprule
System & Operator log-slope & Baseline log-slope \\
\midrule
miR-122   & 0.874 & 0.000 \\
miR-92/25 & 0.674 & 0.000 \\
\bottomrule
\end{tabular}
\caption{Competitive derepression, log-slope separation. Log-slope of predicted reporter derepression in competitor abundance: strictly positive for the operator, exactly zero for every independent per-target scorer, because competitor abundance is not an argument of the scoring function. The zero is a property of the input signature, not of tuning, and is calibration-free.}
\label{tab:derep-tier1}
\end{table}

\subsection{Fixed Constants}
\label{app:constants}

The following constants are fixed inputs, not fitted parameters. None is tuned against the quantity it is used to predict.

\paragraph{Thermodynamic conversion (design validation).} The effect of the (S)-GNA seed modification is entered as a change in the seed dissociation constant derived from a measured melting-temperature shift, using a two-state nearest-neighbor model \citep{santalucia1998}. With $\Delta G(T) = \Delta H\,(1 - T/T_m)$, the modification free-energy change is $\Delta\Delta G = |\Delta H|\,T\,|\Delta T_m| / T_m^2$, evaluated at $\Delta H = -150$ kcal/mol, $T_m = 340$ K, and body temperature $T = 310$ K. At the literature single-modification melting shift $\Delta T_m = 1.5\,^\circ\mathrm{C}$, this gives $\Delta\Delta G \approx 0.60$ kcal/mol and a seed-affinity factor $K_{\text{mod}}/K_{\text{parent}} \approx 2.66$. The melting shift is a measured thermodynamic quantity from the (S)-GNA chemistry literature, not a fit to the off-target fold-change data.

\paragraph{Abundance regimes (derepression contrast).} The copies-per-cell miRNA and target abundances in the competitive-derepression contrast are set to each source study's measured miRNA-to-target regime: an abundant-miRNA, high-spare-capacity regime for miR-122 \citep{denzler2014}, and a low
miRNA-to-target-ratio regime for miR-92/25 \citep{bosson2014}. These stoichiometries are measured inputs that place each system in its reported regime; they are not fit to the derepression response. The absolute affinity scale is nominal, so the operator's exact threshold position is calibration-dependent; we state this explicitly, and note that the capability claim (the independent-scorer sensitivity is identically zero) is calibration-free and holds for any positive affinity scale.

\begin{table*}[t]
\centering
\begin{tabular}{lrrrrr}
\toprule
System & $M$ & Operator threshold & Measured anchor & Common ratio & Per-system ratio \\
\midrule
miR-122   & $1.35\times10^{5}$ & $2.37\times10^{5}$ & $1.50\times10^{5}$ & 1.58 & 1.60 \\
miR-92/25 & $1.30\times10^{4}$ & $4.48\times10^{4}$ & $3.00\times10^{3}$ & 14.9 & -- \\
\bottomrule
\end{tabular}
\caption{Competitive derepression, threshold ordering (added-competing-site axis). Loaded guide abundance $M$, operator-predicted threshold, measured anchor \citep{denzler2014,bosson2014}, and predicted-over-measured ratio under a common calibration-free background (default) and under a per-system measured background. The per-system ratio is reported only for miR-122, whose measured effective target-site abundance is $3\times10^{4}$ \citep{denzler2016}, a different study from the one supplying its threshold and therefore independent of it; for miR-92/25 the only measured background available equals its own threshold, so no independent per-system ratio exists (--). Even for miR-122 the measured background barely moves the prediction (1.58 to 1.60): the calibration-free operator did not need it. The operator recovers the ordering of the two systems but not the magnitude of their separation: the measured thresholds differ about fiftyfold and the predicted ones about fivefold, with miR-92/25 over-predicted fifteenfold. Only the ordering is claimed.}
\label{tab:derep-thr}
\end{table*}

\paragraph{Ordering robustness to the absolute affinity scale.}
The absolute dissociation-constant scale in copies per cell is nominal and cannot be pinned from the
data; only the relative site-type ordering ($8$mer $<$ $7$mer $<$ $6$mer) is fixed. Multiplying that
scale by a factor swept over $[0.1, 100]$ (three orders of magnitude, ten points), with the relative
ordering preserved, the operator's predicted derepression threshold for miR-122 exceeds that for
miR-92/25 at every scale tested. The separation ratio compresses monotonically from $17.6\times$ to
$3.1\times$, driven almost entirely by the miR-92/25 threshold, which rises about sevenfold across
the sweep; the miR-122 threshold moves by a factor of about $1.2$ only, from
$1.59\times10^{5}$ to $1.92\times10^{5}$. The ordering never inverts. Two scope restrictions apply and we state both. This
sweep is computed under a derepression-crossing threshold definition, which is not the definition
tabulated in Table~\ref{tab:derep-thr}; it therefore establishes robustness of the ordering, and we
do not claim it establishes robustness of the tabulated thresholds. And it demonstrates invariance to
the affinity calibration specifically, not to the assumed endogenous copy-number pools, which are
held at each study's measured regime; the margin is scale-dependent even though its sign is not.

\subsection{Parameter Sweeps and Selection Criteria}
\label{app:sweeps}

Where a modeling input is not pinned by a measurement, we report the result across a swept range rather than selecting a single value.

\paragraph{Melting-shift sweep (design validation).} The seed-affinity change is swept over $\Delta T_m \in \{1.0, 1.5, 3.0\}\,^\circ\mathrm{C}$, spanning the plausible range for a single (S)-GNA substitution, giving seed-affinity factors from $\approx 1.9$ to the saturating end. The certified conclusion (the modification is a certified off-target reducer, i.e. the certified occupancy bound decreases) holds across the entire sweep; only the predicted magnitude, not the certified direction, varies with $\Delta T_m$. The central value $1.5\,^\circ\mathrm{C}$ is the literature single-modification estimate.

\paragraph{Submodularity regimes (non-submodularity check).} The submodularity enumeration is repeated across three loading regimes (subsaturating, balanced, saturating) and three modification strengths ($\delta \in \{0.7, 1.5, 3.0\}$), giving nine regime-strength combinations. The verdict (submodular in the isolated model within subsaturated regimes, non-submodular in the coupled model in every regime) is reported for all nine in Table~\ref{tab:submod}, with the maximum diminishing-returns violation per combination. Marginal gains are nonnegative except in the guide-limited subsaturating regime at $\delta\ge1.5$, where a marginal modification can raise total off-target occupancy through the shared on-target sink (Table~\ref{tab:submod}, $^{\dagger}$); the supermodularity verdict is unaffected. The enumeration is exhaustive over all $2^{6}$ subsets and all $1{,}458$ order triples, so there is no sampling and no selection.

\paragraph{Occupancy-form check (form independence).} The certified-sign verification is repeated for three occupancy functional forms (Langmuir; Hill with $h=2$; saturating-exponential) across all $17$ miRNAs, confirming form independence. No form is selected as preferred; all three are reported.

\subsection{Evaluation Metrics and Run Counts}
\label{app:metrics}

The theory verification reports, per miRNA, the maximum relative error between the closed-form comparative static and a forward finite difference; it ranges from about $10^{-7}$ to $10^{-6}$ across the $17$ miRNAs for the Langmuir form, reaches $2.3\times10^{-5}$ for the saturating-exponential form, and is near $10^{-4}$ for the Hill form, where the occupancy is non-concave below its inflection. The forward step is $h = 10^{-5}\max(x_s,1)$ on the largest-abundance coordinate, with equilibrium re-solve tolerance $10^{-14}$ for the Langmuir form and $10^{-13}$ for the Hill and saturating-exponential forms. No magnitude mask is applied: the relative error is $|\mathrm{fd}-\mathrm{cf}|/|\mathrm{cf}|$ over every entry of the sensitivity vector, guarded only against division by exact zero, so each reported maximum is a maximum over the full vector rather than over a filtered subset. The single-pool M-matrix property and certified sign are trivial-by-construction (the Jacobian is diagonal and positive) and are not counted as tests, since they cannot fail; the load-bearing checks are the finite-difference agreement and its form-independence. Because these computations are deterministic, each is a single exhaustive run; the $2^{6}$ enumerated submodularity subsets are likewise exhaustive rather than sampled. The design-principle validation reports the one-sided Kolmogorov-Smirnov statistic $D$ (not the Jacobian diagonal $D_i$ of Table~1 of the main paper) for the left shift of seed-matched genes within each arm (parent $D = 0.118$, modified $D = 0.035$), the attenuation $1 - D_{\mathrm{mod}}/D_{\mathrm{parent}} = 0.71$ on the shared universe of $3{,}206$ seed-matched genes against $16{,}760$ background genes, and a $2000$-sample stratified bootstrap $95\%$ confidence interval on the attenuation ($[0.57, 0.86]$) that is disjoint from the null RISC-block contrast. Attenuation is computed from unrounded $D$: the stored values are $D_{\mathrm{parent}} = 0.117966$ and $D_{\mathrm{mod}} = 0.034537$, giving $0.7072$. The three-decimal figures quoted elsewhere are rounded for display, and recomputing the ratio from those rounded values gives $0.703$, which agrees with the reported attenuation only to one decimal place; we give the unrounded values here so the headline figure is reproducible from printed numbers. Significance of each arm's shift is assessed by the Kolmogorov-Smirnov test against the non-matched background and confirmed by a $2000$-shuffle permutation null; the permutation $p$-values track the analytic KS $p$-values across all arms.

\begin{table}[t]
\centering
\begin{tabular}{lrrrr}
\toprule
microRNA & Targets & $D$ & Max rel.\ err. & Mean derep. \\
\midrule
let-7    & 1175 & $1.59\times10^{4}$ & $1.0\times10^{-6}$ & $-1.07\times10^{-4}$ \\
lsy-6    &  341 & $2.82\times10^{3}$ & $8.5\times10^{-7}$ & $-1.11\times10^{-4}$ \\
miR-1    & 1599 & $1.40\times10^{4}$ & $1.2\times10^{-7}$ & $-2.47\times10^{-5}$ \\
miR-124  & 2198 & $1.75\times10^{4}$ & $1.8\times10^{-7}$ & $-1.92\times10^{-5}$ \\
miR-137  & 1066 & $1.26\times10^{4}$ & $3.6\times10^{-7}$ & $-9.57\times10^{-5}$ \\
miR-139  & 1512 & $1.52\times10^{4}$ & $2.8\times10^{-7}$ & $-6.95\times10^{-5}$ \\
miR-143  & 1776 & $2.06\times10^{4}$ & $5.2\times10^{-7}$ & $-8.13\times10^{-5}$ \\
miR-144  & 1589 & $1.82\times10^{4}$ & $8.4\times10^{-8}$ & $-1.82\times10^{-5}$ \\
miR-153  & 1249 & $9.93\times10^{3}$ & $1.4\times10^{-7}$ & $-3.38\times10^{-5}$ \\
miR-155  & 1052 & $1.16\times10^{4}$ & $3.2\times10^{-7}$ & $-1.00\times10^{-4}$ \\
miR-182  & 1733 & $2.29\times10^{4}$ & $3.3\times10^{-7}$ & $-7.37\times10^{-5}$ \\
miR-199a & 1843 & $1.68\times10^{4}$ & $1.7\times10^{-7}$ & $-1.99\times10^{-5}$ \\
miR-204  & 1767 & $2.67\times10^{4}$ & $9.1\times10^{-7}$ & $-1.38\times10^{-4}$ \\
miR-205  & 1818 & $2.34\times10^{4}$ & $2.1\times10^{-6}$ & $-1.42\times10^{-5}$ \\
miR-216b & 1501 & $1.59\times10^{4}$ & $1.1\times10^{-7}$ & $-2.05\times10^{-5}$ \\
miR-223  & 1281 & $1.32\times10^{4}$ & $2.4\times10^{-7}$ & $-2.48\times10^{-5}$ \\
miR-7    & 2098 & $1.88\times10^{4}$ & $2.7\times10^{-7}$ & $-5.94\times10^{-5}$ \\
\bottomrule
\end{tabular}
\caption{Theory verification on the McGeary corpus (GSE140217), Langmuir occupancy. Per microRNA: seed-matched target count, the residual-Jacobian diagonal $D=1+\sum_s x_s\sigma_s'\ge 1$ (distinct from the Kolmogorov-Smirnov $D$ of Appendices~\ref{app:metrics} and \ref{app:controls}; single-pool $J=D$ is an M-matrix by construction, so this column is informational, not a test), the maximum relative error between the closed-form comparative static and a forward finite-difference re-solve, and the mean equilibrium derepression sensitivity (negative throughout, confirming the certified sign). The closed-form/finite-difference agreement is the load-bearing check; its maximum over all 17 microRNAs is $2.1\times10^{-6}$ (Langmuir), $2.3\times10^{-5}$ (saturating-exponential), and $1.4\times10^{-4}$ (Hill, non-concave below its inflection), confirming form-independence of the sign.}
\label{tab:verify}
\end{table}

\subsection{Trainable-Layer Experiment}
\label{app:training}
We instantiate the conservation-law operator as a trainable layer by learning the per-site
affinity map $K_\theta(\cdot)=\mathrm{softplus}(\mathrm{MLP}(\cdot))+\epsilon$ with a positive
floor $\epsilon=10^{-6}$, so $K_\theta>0$ and assumption (A1) holds by
construction. The MLP consumes $d=7$ site features. The forward pass solves
the single-pool equilibrium by bisection; the backward pass returns the closed-form
implicit-function-theorem gradient rather than differentiating through the solver.
We train on the McGeary GSE140217 corpus with Adam (learning rate $0.01$) for $300$ full-batch steps (no minibatching; the
loss is summed over all 13 training microRNAs and one optimizer step is taken
per epoch), in float64 on CPU with an explicit seed $20240716$ and
deterministic algorithms. The equilibrium layer has 4739 learnable parameters and the same-capacity feedforward baseline has 4737 (within 0.05\%).
Four microRNAs are held out of training and evaluation (\textit{lsy6}, \textit{mir139}, \textit{mir144}, \textit{mir153}); the pre-training gradient
check uses three of them (\textit{lsy6}, \textit{mir139}, \textit{mir144}), so the three gradient-check microRNAs are a subset of the
four held-out microRNAs (all but \textit{mir153}). Across those three the custom backward matches a Richardson-extrapolated central finite difference ($h = 10^{-4}$, fourth order; central here, unlike the forward difference of Appendix~\ref{app:metrics}, because this
is a gradient check on the backward pass rather than on the closed-form
comparative static) to a relative $L_2$ error of at most $2.1\times10^{-7}$
($<3\times10^{-7}$). The certified precondition holds at every step: the minimum single-pool
Jacobian diagonal over all $300$ steps is $101.18>0$.
On held-out data the equilibrium layer and the baseline reach parity in magnitude. Over ten fixed
seeds (split held at the original seed so the four held-out microRNAs are unchanged), the
equilibrium layer averages mean per-microRNA Spearman $0.112$ (sample sd $0.013$, range
$[0.086, 0.122]$) against the baseline's $0.095$ (sd $0.015$, range $[0.069, 0.121]$). The
per-seed ranges overlap and two baseline seeds exceed five equilibrium seeds, so the separation is
within seed variance; the single-seed values reported above ($0.118$ and $0.097$) are one draw. The
certificate held on every step of every seed, minimum single-pool Jacobian diagonal $101.18$ over
all $3{,}000$ steps; the global minimum falls in the original-seed run, whose per-seed minimum is
the lowest of the ten (next lowest $104.00$), which is why the two reported minima coincide. Only
the per-run minimum was logged, so the step index at which it occurs is not recoverable from the
stored artifacts.

\subsection{Contrasts That Cannot Answer the Question}
\label{app:unmeasurable}
We report both rather than omit them. In the alkyl-phosphonate series of \citet{nikan2025}
(GSE285764), the in vivo \emph{Ttr} arms are perfectly confounded with batch: every parent sample is
in one study and every modified sample in another, each with its own vehicle. The certified sign flips
with the control choice (own-batch attenuation $-0.75$, pooled $+0.62$), so no design separates the
modification from the batch.
In the in vitro ACTN1 arms of the same series, the guide seed matches $46\%$ of annotated human
$3'$UTR genes, against $9.4\%$ (Ttr) and $2.0\%$ (Hao1) of annotated rat $3'$UTR genes. Each rate is
computed on its own guide's species annotation, so we read the $46\%$ as an absolute statement about
the ACTN1 matched set rather than as a ratio against the other two. At that rate the union-class
matched set is close to a random sample of the transcriptome and tracks any global shift rather than
a seed-specific one.
Restricting to $8$mer sites gives a specific set, but its attenuation sign flips under two independent
benign perturbations (annotation choice, control-set choice), so it carries no information. The
contrast is unmeasurable in either direction. A third guide in the series is dropped because its
sequence is not published in verifiable form.

\subsection{Scope of the Certificate}
\label{app:scope}
The operator's prediction requires a map from chemistry to a change in seed affinity. For (S)-GNA a
free-duplex thermal destabilization is reported, and the certified sign is invariant to its exact
value: sweeping the affinity ratio $R = K_{\mathrm{mod}}/K_{\mathrm{parent}}$ over $[1.1, 10]$ gives
$\Delta\Phi < 0$ throughout, so no calibration carries the claim. For alkyl-phosphonate backbone
modifications, no measured thermodynamic value establishes an increase in seed $K$: the modification
perturbs backbone geometry in the Argonaute channel rather than base pairing, and
\citet{nikan2025} report that thermal destabilization does not predict off-target mitigation, with
adjacent linkage positions differing roughly twenty-fold at similar $\Delta T_m$. The certificate's
antecedent is therefore not established for that chemistry and the operator makes no prediction there.
We state plainly that this scope restriction was adopted after the alkyl-phosphonate contrasts
returned no usable signal; the mechanistic grounds for it were available beforehand and we give them,
but we do not present the restriction as having been specified in advance.

\subsection{Mechanistic Control and Per-Guide Variation}
\label{app:controls}

\paragraph{Mechanistic control.}
The operator presumes that occupancy is driven by a conserved pool of \emph{loaded} guide. We test
this directly using the RISC-loading-blocked arm of \citet{janas2018} (GSE108823), in which the same
guide carries an inactive backbone that prevents loading. The seed signature vanishes: $D = 0.006$
(Kolmogorov-Smirnov, not the Jacobian diagonal of Table~\ref{tab:verify}), $p = 0.96$, on $631$
seed-matched genes, against $D = 0.147$ ($650$ genes) and $D = 0.180$ ($625$ genes) for the loaded
parent in the study's two sequencing batches. Each arm is scored within its own batch's tested
universe, so we give the parent value from each batch rather than a single comparator; both exceed
the blocked arm by more than an order of magnitude. Loading the guide into RISC is what produces the
occupancy signature the operator models.

\paragraph{The intervention does not work on every guide.}
On a second guide from the same chemistry and the same seed position (Hao1, (S)-GNA at position
seven, GSE108823), the modification barely moves the off-target signature: parent $D = 0.147$
($p = 1.6\times10^{-12}$), modified $D = 0.143$ ($p = 6.2\times10^{-12}$), attenuation $0.025$,
computed from the unrounded values $D_{\mathrm{parent}} = 0.146802$ and
$D_{\mathrm{mod}} = 0.143147$; recomputing from the three-decimal figures gives $0.027$, so we print
the unrounded values here as we do for Ttr. The
modified arm remains as strongly significant as the parent. The certified sign is technically
satisfied, since $0.143 < 0.147$, but the magnitude is indistinguishable from zero. We disclose that
this verdict depends on the $3'$UTR annotation: under a single-isoform-per-gene annotation the
attenuation reads $0.298$, and only the union-over-isoforms annotation (correct, because a gene
carrying a seed site in any isoform belongs in the matched set rather than the background) gives
$0.025$. The conclusion is the honest one: the same chemistry at the same seed position produces an
attenuation of $0.71$ on one guide and effectively none on another, and the single-pool model, which
sees only aggregate occupancy, does not distinguish the two cases. This is the sharpest statement of
the instantiation's magnitude-ranking limitation.

\subsection{A Decision an Independent Scorer Cannot Inform}
\label{app:decision}
The structural blindness of Proposition~2 of the main paper is not only a derivative identity; it
changes concrete design decisions. Consider two conditions for the same miR-122 guide and the same
reporter, differing only in the abundance of a competing transcript: condition A at the measured
effective endogenous target-site abundance ($3\times10^{4}$ sites per cell) and condition B at the
measured half-maximal derepression point ($4.5\times10^{5}$ sites, the IC50, distinct from the
$1.5\times10^{5}$ onset threshold in Table~\ref{tab:derep-thr}), both real anchors from
\citet{denzler2016}. To an independent per-target scorer the two conditions are indistinguishable:
the focal target's sites are identical, the competitor is not an argument of the scoring function,
and so every scorer we tested returns a bit-identical prediction for A and B with
competitor-abundance derivative exactly zero (seed count, seed-pairing stability, an
independent-occupancy score, and a context regression all agree $A = B$). A designer relying on any
of them would treat the two conditions as interchangeable. The operator distinguishes them,
predicting the focal target derepressed to $0.027$ at A (sub-threshold) and $0.711$ at B (above
half-maximal), an ordering $B > A$ that the measured titration confirms, since B sits at the measured
IC50 and A lies below the measured release threshold. We are precise about what this does and does
not show. The operator's derepression is monotone in competitor abundance by the M-matrix property
itself, so it necessarily orders B above A; the measured titration is also monotone, so the two
cannot disagree here, and this is not a falsification test of the operator. What it demonstrates is
the decision-relevance of the capability: the ordering is one the operator and the measured biology
both resolve and that no independent scorer can express for any calibration. As with the other
instantiations, the magnitude is over-predicted ($0.71$ versus the measured $0.5$) and only the
ordering is claimed.

\subsection{Summary of Reproducibility-Relevant Choices}
\label{app:reprosummary}

Every load-bearing theoretical result (the M-matrix characterization, the phase-boundary sign condition, and the non-submodularity result) is deterministic and verified either by exhaustive enumeration or by closed-form-versus-numerical agreement, so reproduction requires no seed control and no hyperparameter tuning. The two places where a modeling input is not fixed by a measurement (the melting-shift-to-affinity conversion and the abundance calibration in the derepression contrast) are reported across swept ranges, with the certified qualitative conclusion shown to be invariant across the sweep and only the predicted magnitude varying. The absolute affinity scale in the derepression contrast is nominal and its effect on the operator's exact threshold is stated as a limitation, while the calibration-free capability claim is separated out explicitly.

\end{document}